%% file: main.tex
\documentclass[10pt,twocolumn,letterpaper]{article}

\usepackage[pagenumbers]{cvpr} 

\input{preamble}

\usepackage{graphicx}
\usepackage{subcaption}
\usepackage{amsmath}
\usepackage{amssymb}
\usepackage{booktabs}

\usepackage{url}            
\usepackage{booktabs}       
\usepackage{amsfonts}       
\usepackage{nicefrac}       
\usepackage{microtype}      
\usepackage{xcolor}         

\usepackage{diagbox}
\usepackage[ruled, vlined]{algorithm2e}%
\SetKwInOut{Parameter}{parameter}
\usepackage{wrapfig}
\usepackage{enumitem,amsthm,amsmath,amsfonts,amssymb}
\usepackage{caption}
\usepackage{subcaption}
\usepackage{multirow}
\usepackage{paralist}
\usepackage{makecell}
\usepackage{color}
\usepackage{xcolor}
\usepackage{threeparttable}
\usepackage{amsfonts}
\usepackage{algorithmicx}

\newtheorem{theorem}{Theorem}

\newcommand{\indep}{\rotatebox[origin=c]{90}{$\models$}}

\usepackage{dsfont}

\def\sign{\texttt{sign}}

\def\1{\cellcolor{green!30}}
\def\2{\cellcolor{green!30}}
\def\3{\cellcolor{yellow!30}}
\def\4{\cellcolor{yellow!30}}
\def\5{\cellcolor{yellow!30}}
\def\6{\cellcolor{orange!40}}
\def\7{\cellcolor{orange!40}}
\def\8{\cellcolor{orange!40}}
\def\9{\cellcolor{orange!40}}

\definecolor{color_best}{RGB}{22, 138, 173}

\usepackage{tikz}

\usepackage{colortbl}

\definecolor{baselinecolor}{gray}{.9}
\newcommand{\baseline}[1]{\cellcolor{baselinecolor}{#1}}

\definecolor{cvprblue}{rgb}{0.21,0.49,0.74}
\usepackage[pagebackref,breaklinks,colorlinks,citecolor=cvprblue]{hyperref}

\def\paperID{2840} 
\def\confName{CVPR}
\def\confYear{2024}

\title{Preserving Fairness Generalization in Deepfake Detection\thanks{\textcolor{blue}{This paper has been accepted by CVPR 2024}}}

\author{Li Lin$^{1}$, Xinan He$^{2}$, Yan Ju$^{3}$, Xin Wang$^{4}$, Feng Ding$^{2}$, Shu Hu$^{1}$\thanks{Corresponding author}\\
$^{1}$Purdue University {\tt \small\{lin1785, hu968\}@purdue.edu}\\
$^{2}$Nanchang University {\tt \small\{shahur, fengding\}@ncu.edu.cn}\\
$^{3}$University at Buffalo, State University of New York \tt \small yanju@buffalo.edu\\
$^{4}$University at Albany, State University of New York \tt \small xwang56@albany.edu\\
}

\begin{document}

\maketitle
\begin{abstract}
Although effective deepfake detection models have been developed in recent years, recent studies have revealed that these models can result in unfair performance disparities among demographic groups, such as race and gender. This can lead to particular groups facing unfair targeting or exclusion from detection, potentially allowing misclassified deepfakes to manipulate public opinion and undermine trust in the model.
The existing method for addressing this problem is providing a fair loss function. It shows good fairness performance for intra-domain evaluation but does not maintain fairness for cross-domain testing. This highlights the significance of fairness generalization in the fight against deepfakes. 
In this work, we propose the first method to address the fairness generalization problem in deepfake detection by simultaneously considering features, loss, and optimization aspects. Our method employs disentanglement learning to extract demographic and domain-agnostic forgery features, fusing them to encourage fair learning across a flattened loss landscape.  Extensive experiments on prominent deepfake datasets demonstrate our method's effectiveness, surpassing state-of-the-art approaches in preserving fairness during cross-domain deepfake detection. The code is available at \url{https://github.com/Purdue-M2/Fairness-Generalization}.
 
\end{abstract}

\vspace{-2mm}
\section{Introduction}
\vspace{-2mm}



Deepfakes, a portmanteau of ``deep learning'' and ``fake,'' have emerged as a captivating yet concerning facet of contemporary technology. These are AI-generated or manipulated media (\eg, images, videos) through deep neural networks (\eg, variational autoencoder \cite{vahdat2020nvae},  generative adversarial networks \cite{karras2020analyzing}, diffusion models \cite{dhariwal2021diffusion}) that appear startlingly genuine, often featuring individuals engaged in actions they never partook in or uttering words they never spoke. 
While deepfakes have opened doors to creative content and entertainment, malicious use of deepfakes can lead to misinformation, privacy breaches, and even political manipulation, eroding trust and generating confusion \cite{wang2022gan, masood2023deepfakes}.

\begin{figure}[t]
  \centering
  \includegraphics[width=1\linewidth]{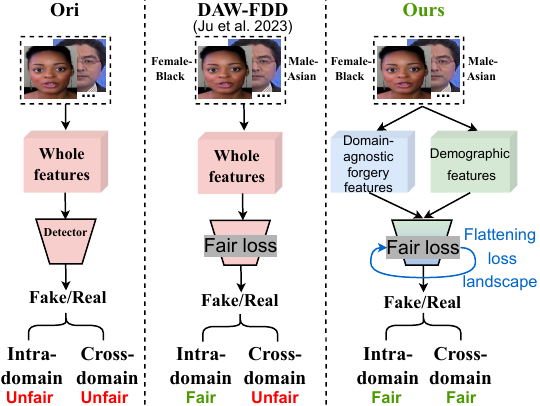}
  \caption{\small Comparison between our method and existing deepfake detection baselines. (\textbf{Left}) The Ori represents the conventional method without any fair characters. (\textbf{Middle}) The DAW-FDD \cite{ju2023improving} is an intra-domain fair deepfake detection method. However, this method fails in cross-domain fair detection. 
  (\textbf{Right}) Our method succeeds in achieving both intra-domain and cross-domain fair detection by exposing domain-agnostic forgery features and demographic features and then fusing them for fair learning across a flattened loss landscape. }
  \vspace{-5mm}
  \label{fig:introduction}
\end{figure}

To counteract the spread of deceptive deepfakes, there is a burgeoning field of deepfake detection methods that are data-driven and deep-learning based \cite{marra2019incremental, goebel2020detection, wang2020cnn, liu2020global, hulzebosch2020detecting, guo2022robust, pu2022learning, hu2021improving, zhang2020face, liang2022exploring,Yan_2023_ICCV,zheng2024few, chen2024masked, lin2024detecting, fan2023synthesizing, zhang2023x, yang2023improving, fan2023attacking, chen2023harnessing}. 
However, recent research and reports \cite{trinh2021examination, xu2022comprehensive, news2, nadimpalli2022gbdf, masood2022deepfakes} have brought to light fairness issues within current deepfake detection methods. One significant concern revolves around the inconsistency in performance when assessing different demographic groups, including gender, age, and ethnicity~\cite{xu2022comprehensive}. For example, some of the most advanced detectors exhibit higher accuracy when evaluating deepfakes featuring individuals with lighter skin tones compared to those with darker skin tones~\cite{hazirbas2021towards, trinh2021examination}. This allows attackers to generate harmful deepfakes targeting specific populations in order to evade detection.  

An initial algorithm-level approach to addressing fairness in deepfake detection has been presented by Ju et al. \cite{ju2023improving}. 
They showed that the proposed DAW-FDD model could exhibit the best fairness performance under the intra-domain evaluation scenario, \ie, training and testing data are generated by the same forgery techniques. However, in practice, we found that their method does not preserve fairness for cross-domain evaluation, \ie, when testing on data generated by unknown forgeries. 
Notably, achieving fairness generalization is critical. 
Without such generalization, the current fair deepfake detection methods are susceptible to obsolescence easily.   

In this work, we experimentally and theoretically analyze the entanglement of demographic and forgery features, and the sharpness of loss landscapes could be the fuse to affect the fairness generalization in deepfake detection. 
To address these issues, we propose a novel framework to preserve fairness in deepfake detection generalization, consisting of three key modules: disentanglement learning, fairness learning, and optimization. 
Specifically, in the disentanglement learning module, we introduce a disentanglement loss to expose demographic and domain-agnostic forgery features –– the feature-level factors directly affecting the fairness generalization capabilities of the detector.  The fairness learning module combines these disentangled features to promote fair learning while guided by generalization principles. Additionally, we include a bi-level fairness loss to enhance fairness both across and within subgroups.  The optimization module focuses on flattening the loss landscape, allowing the model to escape suboptimal solutions and fortify its fairness generalization capability. Fig.~\ref{fig:introduction}  illustrates how our method differs from existing ones.
Our contributions are as follows:
\begin{compactitem}
    \item We experimentally and theoretically analyze the unfairness problem in deepfake detection generalization.
    \item We propose the first method to improve fairness generalization in deepfake detection by simultaneously addressing features, loss, and optimization. Specifically, we utilize disentanglement learning to extract demographic and domain-agnostic forgery features, which are then integrated to facilitate fair learning across a flattened loss landscape.    
    \item Our method outperforms state-of-the-art approaches in preserving fairness during cross-domain deepfake detection, as demonstrated in extensive experiments on various leading deepfake datasets. 
\end{compactitem}

\section{Related Work}
%
\textbf{Deepfake Detection}.  
The largest portion of existing deepfake detection methods fall into the \textit{data-driven} category, including \cite{marra2019incremental, goebel2020detection, wang2020cnn, liu2020global, hulzebosch2020detecting, guo2022robust, pu2022learning}. These methods leverage various types of Deep Neural Networks (DNNs) trained on both authentic and deepfake videos to capture specific discernible artifacts. While these methods have achieved promising performance for the intra-domain evaluation, they suffer from sharp performance degradation on cross-domain testing. 
To address the generalization issue, disentanglement learning \cite{wang2022disentangled} is widely used for forgery detection by extracting relevant features while eliminating irrelevant ones. For instance, Hu et al. \cite{hu2021improving} introduced a disentanglement framework to automatically locate forgery-related regions, and Zhang et al. \cite{zhang2020face} enhanced generalization through auxiliary supervision. Liang et al. \cite{liang2022exploring} proposed a framework that improves feature independence through content consistency and global representation contrastive constraints. Yan et al. \cite{Yan_2023_ICCV} extended this framework by exclusively utilizing common forgery features, which are separated from forgery-related features.

\begin{figure*}[t]
    \centering
    \begin{minipage}{0.7\textwidth}
        \centering
        \includegraphics[width=\linewidth]{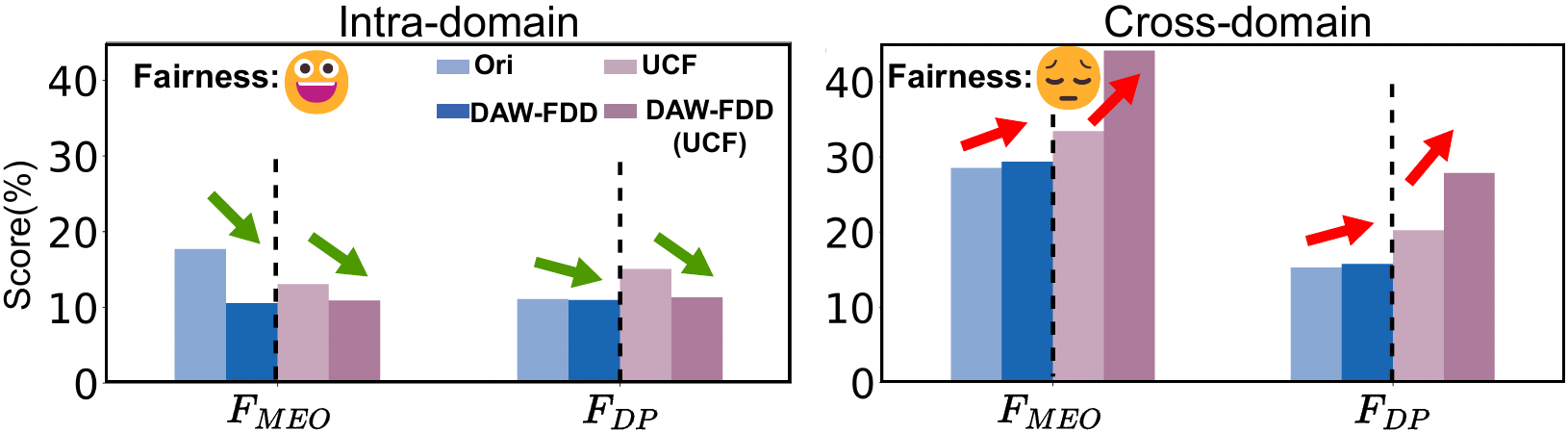}
    \end{minipage}\hspace{0.5cm}
    \begin{minipage}{0.25\textwidth}
        \centering
        \includegraphics[width=\linewidth,height=0.8\textwidth]{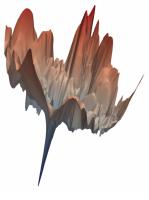}
    \end{minipage}
    \vspace{-2mm}
    \caption{\small Experimental results for Motivation. 
    Testing fairness results (lower is better for all metrics) of deepfake detectors in intra-domain (\textbf{Left}, train and test: FF++) and cross-domain (\textbf{Middle}, train: FF++, test: DFD) detection. 
    (\textbf{Right}) Visualization of loss landscape for DAW-FDD. 
    The numerous local and global minima could cause the model to have poor generalization. 
    }    
    \label{fig:motivation}
\vspace{-5mm}
\end{figure*}

\smallskip
\noindent\textbf{Fairness in Deepfake Detection}. Recent studies have mentioned fairness issues in deepfake detection \cite{masood2022deepfakes}. Trinh et al. \cite{trinh2021examination} identified biases in both deepfake datasets and detection models, revealing significant error rate differences across subgroups. Similar observations were reported in the study by Hazirbas et al. \cite{hazirbas2021towards}. Pu et al. \cite{pu2022fairness} assessed the fairness of the MesoInception-4 deepfake detection model on FF++ and found it to be unfair to both genders. Xu et al. \cite{xu2022comprehensive} conducted a comprehensive analysis of bias in deepfake detection, enriching datasets with diverse annotations to support future research. Additionally, Nadimpalli et al. \cite{nadimpalli2022gbdf} highlighted substantial bias in datasets and detection models, introducing a gender-balanced dataset to mitigate gender-based performance bias. However, this approach yielded only modest improvements and required extensive data annotation. Ju et al. \cite{ju2023improving} focused on enhancing fairness within the same data domain but did not address fairness in cross-domain testing, which is the central focus of our paper.



\section{Motivation} 
\label{sec:motivationshort}


\textbf{Unfairness in Cross-domain Detection}. 
To assess the performance of existing fair deepfake detection methods in ensuring fairness across different testing domains, we utilized the DAW-FDD method \cite{ju2023improving} with an Xception backbone. For comparison, we employed a baseline detector with the same backbone and cross-entropy loss, and named it `Ori'. To evaluate the effectiveness of incorporating fairness loss in generalized detectors, we examined the UCF baseline \cite{Yan_2023_ICCV} and trained it with the DAW-FDD fair loss during training, denoted as DAW-FDD (UCF).
All models were trained on the FF++ dataset \cite{rossler2019faceforensics++} and were subsequently tested on both the FF++ and DFD \cite{googledeepfakes2019} datasets. Fairness performance was assessed in terms of demographic group intersection using two fairness metrics: $F_{MEO}$ \cite{wang2023aleatoric} and $F_{DP}$ \cite{wang2022understanding} (details provided in Appendix~\ref{appendix:fairness_metrics}).


The comparison results are presented in Fig.~\ref{fig:motivation} (Left \& Middle). 
The intra-domain testing results reveal that the fairness scores of DAW-FDD and DAW-FDD (UCF) are consistently lower across all metrics when compared to Ori and UCF, respectively. However, in cross-domain testing, DAW-FDD's fairness scores are worse than those of Ori, highlighting the challenge of maintaining fairness when applied across different domains. Additionally, DAW-FDD (UCF) has fairness scores worse than UCF, indicating that merely integrating a fair loss into generalized deepfake detectors is insufficient to ensure successful fairness generalization in cross-domain scenarios.


\smallskip
\noindent
\textbf{Analysis}.
Next, we investigate why current methods fall short in preserving fairness in cross-domain detection, examining both features and optimization-related aspects. In this analysis, we use variables: $X$ (\emph{e.g.}, an image), $Y$ (the corresponding target variable, \emph{e.g.}, fake or real), $\hat{Y}$ (the classifier's prediction for $X$), and $D$ (the demographic variable linked to $X$). Here, $D \in\mathcal{J}$, where $\mathcal{J}$ represents user-defined subgroups (\emph{e.g.}, $\mathcal{J}=$\{male, female\} for gender). For simplicity, we assume $\mathcal{J}$ contains two subgroups, $\mathcal{J}_1$ and $\mathcal{J}_2$.

\smallskip
\noindent
\underline{\textit{Feature Aspect}}. We introduce a theorem as follows:
\begin{theorem}\label{theorem1}
(\cite{locatello2019fairness})
If $X$ is entangled with $Y$ and $D$, the use of a perfect classifier for $\hat{Y}$, \ie, $P(\hat{Y}|X)=P(Y|X)$, does \underline{\textbf{not}} imply demographic parity, \ie, $P(\hat{Y}=y|D=\mathcal{J}_1)=P(\hat{Y}=y|D=\mathcal{J}_2)$, $\forall y\in\{0,1\}$, where 0 means real and 1 means fake.   
\vspace{-2mm}
\end{theorem}
Theorem \ref{theorem1} highlights the challenge of achieving fairness in a model that directly operates on entangled representations $r(X)$ (\ie., $r(X)=X$ when the representations are the
identity function), where these representations are a blend of target information $r(X)_Y$ (for identifying label $Y$) and demographic information $r(X)_D$ (for identifying $D$). This observation suggests a possible reason for the limited success of DAW-FDD \cite{ju2023improving} in fairness generalization.

Therefore, disentanglement could be an approach to enhance fairness by untangling the representations $r(X)_Y$ and $r(X)_D$ from $r(X)$, ensuring their independence, \ie, $r(X)_Y\indep r(X)_D$. Previous methods \cite{hu2021improving, zhang2020face, liang2022exploring, Yan_2023_ICCV} have explored disentanglement learning, particularly in extracting forgery-related features to enhance the generalization of deepfake detection. However, none of these methods address the disentanglement of demographic representation $r(X)_D$. This omission explains why directly applying DAW-FDD to these existing generalization-based models does not preserve fairness in cross-dataset testing. Yet, isolating $r(X)_Y\indep r(X)_D$ could compromise the detection performance of models that rely solely on $r(X)_Y$. This is because forgery and demographic features in deepfakes are often linked to facial characteristics. Removing $r(X)_D$ would result in the loss of facial information that could be related to forgery, potentially causing performance degradation. Hence, this presents a complex challenge that requires careful consideration.

\smallskip
\noindent
 \underline{\textit{Optimization Aspect}}. In addition, existing DNN-based deepfake detection models are highly overparameterized, enabling them to memorize both data and demographic patterns during training. Consequently, the straightforward minimization of commonly used fairness loss functions, such as in the DAW-FDD method, is insufficient to ensure robust fairness generalization. Training these models results in sharp loss landscapes characterized by multiple local and global minima \cite{foret2020sharpness}, each leading to models with varying generalization capabilities due to being trapped into different suboptimal minima. Refer to Fig.~\ref{fig:motivation} (Right) for an example of the DAW-FDD loss landscape. Hence, it becomes essential to flatten the loss landscape to enhance fairness generalization.

\section{Method}
\subsection{Overview of Proposed Method}
According to the insights from Section \ref{sec:motivationshort}, we propose a new method to preserve fairness generalization in deepfake detection in this section. We first formulate the problem.
\begin{figure*}[t]
    \centering
    \includegraphics[width=1\textwidth]{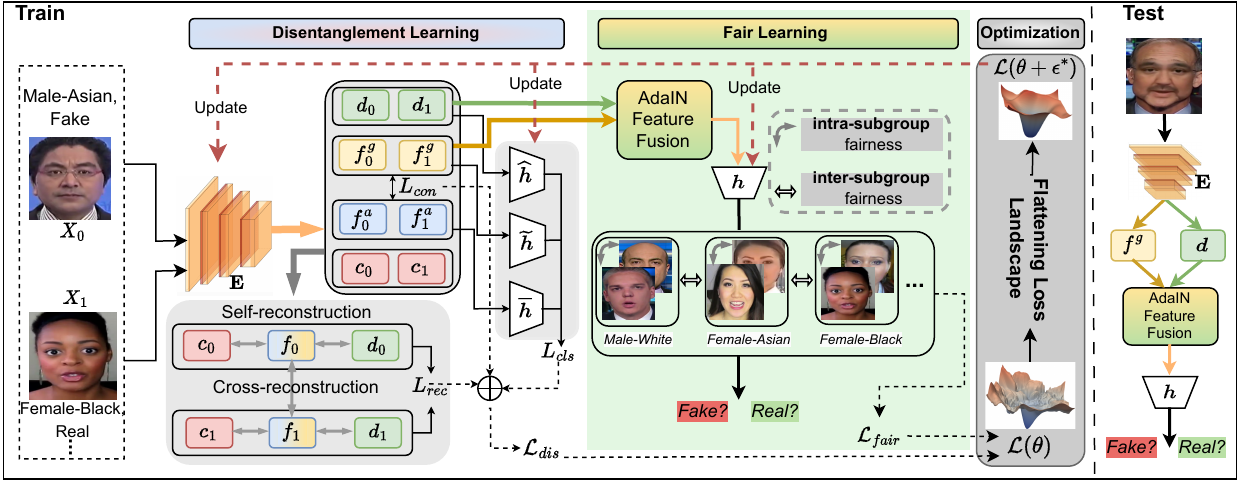}
    \vspace{-6mm}
    \caption{\small An overview of our proposed method. 1) For the disentanglement learning module, we utilize it to expose demographic and forgery features. 2) For the fair learning module, we fuse those two features for a fair classifier head $h$ and obtain the fair prediction using two-level fairness loss $\mathcal{L}_{fair}$. 3) For the optimization module, we flatten the loss landscape to further enhance fairness generalization. }
    \label{fig:overview}
    \vspace{-4mm}
\end{figure*}

\smallskip
\noindent
\textbf{Problem Setup}. Given a training dataset $\mathcal{S}=\{(X_i, D_i, A_i, Y_i)\}_{i=1}^n$ with size $n$. 
$A_i$ represents the domain label, indicating the source of $X_i$. For example,
in the FF++ dataset \cite{rossler2019faceforensics++}, $A_i\in$$\{$\text{real}, \text{DeepFakes} \cite{DeepFakes2017}, \text{Face2Face} \cite{thies2016face2face}, \text{FaceSwap} \cite{FaceSwap2018}, \text{NeuralTextures} \cite{thies2019deferred}, \text{FaceShifter} \cite{li2019faceshifter}$\}$, which correspond to real and fake images generated by various face manipulation methods. Our objective is to train a fair deepfake detection model using $\mathcal{S}$ that can then generalize to an unseen deepfake dataset while maintaining both accuracy and fairness.

\smallskip
\noindent
\textbf{Framework}. Fig.~\ref{fig:overview} depicts our framework, comprising three modules: disentanglement learning, fair learning, and optimization. The disentanglement learning module's purpose is to extract domain-agnostic forgery and demographic features from input images. The fair learning module leverages these two types of features to develop a fair classifier. Both learning modules are supervised by an optimization module, enhancing fairness generalization during model training. We will delve into each module's specifics in the following sections. The entire training process is end-to-end.

\subsection{Exposing Demographic \& Forgery Features}
We propose a disentanglement learning module to extract both demographic features (for fairness) and domain-agnostic forgery features (for generalization). To achieve this, we use pairs of images ($X_i, X_{i'}$), where $X_i$ is fake (or real), $X_{i'}$ is real (or fake), $i,i'\in\{1,\cdots,n\}$, and $i\neq i'$. Each image is processed by an encoder $\mathbf{E}(\cdot)$, which includes three distinct encoders\footnote{The three encoders share the same architecture but with different parameters, and the architecture details can be found in Appendix \ref{appendix:network_details}. } responsible for extracting content features $c$ (\ie, related to the image background), forgery features $f$, and demographic features $d$. Note that the forgery features encompass both domain-specific forgery features $f^a$ (\ie, specific to the forgery method) and domain-agnostic forgery features $f^g$ (\ie, common to various forgery methods). The procedure is formulated as follows,
\vspace{-2mm}
\begin{equation*}
    \begin{aligned}
        c_i, f_i^a, f_i^g, d_i=\mathbf{E}(X_i).
    \end{aligned}
\vspace{-2mm}
\end{equation*}
\smallskip
\noindent
\underline{\textit{Classification Loss}}.
Disentangling domain-specific forgery, domain-agnostic forgery, and demographic features typically involves using cross-entropy (CE) loss for each of them. However, deepfake datasets often suffer from imbalances in demographic subgroup distributions, a fundamental issue in achieving fairness in detection \cite{nadimpalli2022gbdf, mathews2023explainable}. Additionally, conventional CE loss training tends to lead to overfitting on examples from the majority subgroups \cite{wang2019symmetric}, making it unsuitable for learning fair demographic feature representations. To address these challenges, we propose a demographic distribution-aware margin loss inspired by \cite{cao2019learning} as follows:
\vspace{-2mm}
\begin{equation*}
    \begin{aligned}
        M(\widehat{h}(d_i),D_i) = -\log\frac{e^{\widehat{h}^{D_i}(d_i)-\Delta^{D_i}}}{e^{\widehat{h}^{D_i}(d_i)-\Delta^{D_i}}+\sum_{p\neq D_i}e^{\widehat{h}^{p}(d_i)}},
    \end{aligned}
\end{equation*}
where $\Delta^{p}= \frac{\delta}{n_p^{1/4}}$ is a demographic subgroup-dependent margin for $p\in\mathcal{J}$ and $\delta$ is a constant. $n_p$ denotes the number of training data points from subgroup $p$. $\widehat{h}$ is the classification head for $d_i$ and $\widehat{h}^{p}$ represents the output for $p$. 

By incorporating this margin loss, we improve generalization for minority subgroups with small $n_p$ by using larger margins $\Delta^{p}$, promoting unbiased demographic feature representation.
Hence, the total classification loss is:
\vspace{-1mm}
\begin{equation*}
    \begin{aligned}        L_{cls}=C(\widetilde{h}(f_i^g),Y_i)+\rho_1C(\overline{h}(f_i^a),A_i) +\rho_2 M(\widehat{h}(d_i),D_i),
    \end{aligned}
\vspace{-1mm}
\end{equation*}
where $C(\cdot,\cdot)$ is the CE loss. $\overline{h}$ and $\widetilde{h}$ are the classification heads for $f_i^a$ and $f_i^g$, respectively\footnote{These classification heads share the same multilayer perceptron (MLP) architecture but with different parameters.}. $\rho_1$ and $\rho_2$ are two trade-off hyperparameters. Training with the above classification loss enables the encoder to acquire specific feature information, enhancing the model's generalization capability.

\smallskip
\noindent
\underline{\textit{Contrastive Loss}}.
The classification loss, which focuses on individual images, overlooks the image correlations that play a crucial role in enhancing the encoder's representation capabilities. Inspired by contrastive learning \cite{oord2018representation,Yan_2023_ICCV}, we can introduce a contrastive loss to address this gap:
\vspace{-1mm}
\begin{equation*}
    \begin{aligned}        
    L_{con}=[b+\|f_{\text{anchor}}-f_+\|_2-\|f_{\text{anchor}}-f_-\|_2]_+,
    \end{aligned}
\vspace{-1mm}
\end{equation*}
where $f_{\text{anchor}}$ represents anchor forgery features of an image, and $f_+$ and $f_-$ represent its positive counterpart from the same source and the negative counterpart from a different source, respectively.  $b$ is a hyperparameter and $[\cdot]_+=\max\{0,\cdot\}$ is a hinge function. We employ $L_{con}$ for both domain-specific and domain-agnostic forgery features in practice.  For domain-specific forgery features, the source is considered the forgery domain, and the contrastive loss motivates the encoder to learn specific forgery representations. For domain-agnostic forgery features, the source can be either real or fake, and the loss encourages the encoder to learn a generalizable representation that is not tied to any specific forgery method.    

\smallskip
\noindent
\underline{\textit{Reconstruction Loss}}.
To preserve the completeness of the extracted features and maintain consistency between the original and reconstructed images at the pixel level, we employ a reconstruction loss. It is formulated as:
\vspace{-2mm}
\begin{equation*}
    \begin{aligned}        
    &L_{rec}=\|X_i-\mathbf{D}(c_i, f_i, d_i)\|_1+\|X_i-\mathbf{D}(c_i, f_{i'}, d_i)\|_1,
    \end{aligned}
    \vspace{-1mm}
\end{equation*}
where $\mathbf{D}(\cdot,\cdot,\cdot)$ is the decoder responsible for reconstructing an image using the disentangled feature representations (refer to Appendix \ref{appendix:network_details} for architecture details). 
In $L_{rec}$ loss, the first term is the self-reconstruction loss, which minimizes reconstruction errors using the latent features of the input image. 
The second term is the cross-reconstruction loss, which penalizes reconstruction errors by incorporating the partner's forgery feature. These two terms work together to improve feature disentanglement. 

\smallskip
\noindent
\textbf{Disentanglement Loss}.
Therefore, the disentanglement loss for exposing demographic and forgery features is
\vspace{-2mm}
\begin{equation}
    \begin{aligned}        
    &\mathcal{L}_{dis}=\frac{1}{n}\sum_i[ L_{cls}+\rho_3 L_{con} +\rho_4 L_{rec}],
    \end{aligned}
\label{eq:dis}
\vspace{-3mm}
\end{equation}
where $\rho_3$ and $\rho_4$ are trade-off hyperparameters.


\subsection{Fair Learning under Generalization}
Once we acquire both the domain-agnostic forgery features and demographic features, we combine them for the purpose of fairness learning using Adaptive Instance Normalization (AdaIN) \cite{huang2017arbitrary}. The fused feature $I_i$ can be formed as follows,
\vspace{-2mm}
\begin{equation*}
    \begin{aligned}        
    I_i=\sigma(d_i)\Big(\frac{f_i^g - \mu(f_i^g)}{\sigma(f_i^g)}\Big)+\mu(d_i),
    \end{aligned}
\vspace{-1mm}
\end{equation*}
where $\mu(\cdot)$ and $\sigma(\cdot)$ compute the mean and standard deviation of the input feature across spatial dimensions independently for each channel. The combination is necessary because deepfake forgery methods often modify the facial region of an image, which contains essential features for determining demographic information. Ignoring either of these features would significantly reduce fairness generalization performance. Our experiments in Section~\ref{subsec:ablation_study} confirm this.



\smallskip
\noindent
\textbf{Fairness Loss}.
Traditional approaches for achieving fair learning, such as \cite{wang2022understanding, wang2023aleatoric}, often involve adding a fairness penalty to the learning objective. However, these methods can only ensure fairness on specific fairness measures, like demographic parity \cite{dwork2012fairness} or equalized odds \cite{hardt2016equality}, which limits the model's fairness scalability and its ability to work with new datasets. Additionally, even if the overall deepfake dataset has balanced fake and real examples, imbalances can still exist within demographic subgroups, potentially leading to biased learning within those subgroups. 

To address these problems, inspired by \cite{ju2023improving, hu2023rank, hu2022distributionally,hu2022sum,hu2021tkml,hu2020learning,hu2023outlier}, we introduce a bi-level fairness loss as follows:
\vspace{-2mm}
\begin{subequations}\small
    \begin{align}        
    &\mathcal{L}_{fair}=\min_{\eta\in \mathbb{R}}\eta+\frac{1}{\alpha|\mathcal{J}|}\sum_{j=1}^{|\mathcal{J}|}[L_{j}-\eta]_+, \label{eq:fairness1}\\
    &\text{s.t.}\ L_j = \min_{\eta_j\in\mathbb{R}} \eta_j+\frac{1}{\alpha' |\mathcal{J}_j|}\sum_{i:D_i=\mathcal{J}_j}[C(h(I_i), Y_i)-\eta_j]_+.
    \label{eq:fairness2}
    \end{align}
\label{eq:fairness}
\end{subequations}
Here, $|\mathcal{J}|$ represents the size of set $\mathcal{J}$, with each subgroup $\mathcal{J}_j\in\mathcal{J}$, and $|\mathcal{J}_j|$ represents the number of training examples in $\mathcal{J}_j$. $h$ is the classification head for $I_i$, sharing the same MLP architecture as other heads, and $\alpha, \alpha'\in(0,1)$ are two hyperparameters. 
The outer-level formulation (Eq.~(\ref{eq:fairness1})) draws inspiration from the fairness risk measure \cite{williamson2019fairness}, aiming to promote fairness among \textit{inter-subgroups}. The inner-level formulation (Eq.~(\ref{eq:fairness2})) is inspired by distributionally robust optimization (\ie, Conditional Value-at-Risk \cite{levy2020large}), which enhances fairness across both real and fake examples within \textit{intra-subgroup}, thereby bolstering model robustness. 

\subsection{Joint Optimization}

Lastly, we jointly optimize the above two modules in a unified framework. To avoid numerous sharp and narrow minima described in Fig.~\ref{fig:motivation}, we utilize the sharpness-aware minimization method \cite{foret2020sharpness} to flatten the loss landscape. 
Specifically, denoting the model weights of the whole framework as $\theta$, flattening is attained by determining an optimal  $\epsilon^*$ for perturbing $\theta$ to maximize the loss, defined as:
\vspace{-2mm}
\begin{equation}
    \begin{aligned}        \epsilon^*&=\arg\max_{\|\epsilon\|_2\leq \gamma}\underbrace{(\mathcal{L}_{dis}+\lambda\mathcal{L}_{fair})}_{\mathcal{L}}\textbf{(}\theta+\epsilon\textbf{)}\\
    &\approx\arg\max_{\|\epsilon\|_2\leq \gamma}\epsilon^\top\nabla_\theta \mathcal{L}=\gamma\sign(\nabla_\theta \mathcal{L}),
    \end{aligned}
\label{eq:epsion_star}
\vspace{-2mm}
\end{equation}
where $\gamma$ is a hyperparameter that controls the perturbation magnitude, and $\lambda$ is a trade-off hyperparameter. The approximation is obtained using first-order Taylor expansion with the assumption that  $\epsilon$ is small.
The final equation is obtained by solving a dual norm problem, where $\sign$ represents a sign function and $\nabla_\theta \mathcal{L}$ being the gradient of $\mathcal{L}$ with respect to $\theta$. As a result, the model weights are updated by solving the following problem:
\vspace{-2mm}
\begin{equation}
    \begin{aligned}        
    \min_\theta \mathcal{L}\textbf{(}\theta+\epsilon^*\textbf{)}.
    \end{aligned}
\label{eq:sharpness}
\vspace{-2mm}
\end{equation}
The intuition is that the perturbation along the gradient norm direction increases the loss value significantly and then makes the model more generalizable in terms of fairness.  

\smallskip
\noindent
\textbf{End-to-end Training}.
In practice, we first initialize the model weights $\theta$ and then randomly select a mini-batch set $\mathcal{S}_b$ from $\mathcal{S}$, performing the following steps for each iteration on $\mathcal{S}_b$ (see Appendix~\ref{appendix:algorithm_details} for more details about Algorithm):
\begin{compactitem}
    \item Fix $\theta$ and use binary search to find the global optimum of $\eta_j$ since (\ref{eq:fairness2}) is convex w.r.t. $\eta_j$.
    \item Take $L_j$ into (\ref{eq:fairness1}) and use binary search to find the global optimum of $\eta$ since (\ref{eq:fairness1}) is convex w.r.t. $\eta$.
    \item Fix $\eta_j$ and $\eta$, compute $\epsilon^*$ based on Eq.~(\ref{eq:epsion_star}).
    \item Update $\theta$ based on the gradient approximation for (\ref{eq:sharpness}): $\theta\leftarrow\theta-\beta \nabla_\theta \mathcal{L}\big|_{\theta+\epsilon^*}$, where $\beta$ is a learning rate.
    
\end{compactitem}

\section{Experiment}


\subsection{Experimental Settings}
\textbf{Datasets}.
To validate the fairness generalization ability of our proposed method, we train our model on the most widely used benchmark FaceForensics++(FF++) \cite{rossler2019faceforensics++} and test it on FF++, DeepfakeDetection (DFD) \cite{googledeepfakes2019}, Deepfake Detection Challenge (DFDC) \cite{deepfakedetection2021}, and Celeb-DF \cite{li2020celebdf}.  The forged images we use in FF++ are generated by five face manipulation algorithms, including DeepFakes (DF) \cite{DeepFakes2017}, Face2Face (F2F) \cite{thies2016face2face}, FaceSwap (FS) \cite{FaceSwap2018}, NerualTexture (NT) \cite{thies2019deferred}, and FaceShifter (FST) \cite{li2019faceshifter}. Since the original datasets do not have the demographic information of each video or image, we follow Ju et al. \cite{ju2023improving} for data processing, data annotation, and sensitive attributes combination (Intersection). Therefore, the Intersection group contains Male-Asian (M-A), Male-White (M-W), Male-Black (M-B), Male-Others (M-O), Female-Asian (F-A), Female-White (F-W), Female-Black (F-B), and Female-Others (F-O). 
Details of each annotated dataset are in Appendix \ref{appendix:additional_settings}.

\smallskip
\noindent
\textbf{Evaluation Metrics}\label{eval_metrics}.
For detection comparison, the Area Under Curve (AUC) is used to benchmark our approach against previous works, which aligns with the detection evaluation approach adopted in precedent works \cite{Yan_2023_ICCV, luo2021generalizing}. Regarding fairness, we use four distinct fairness metrics to evaluate the effectiveness of our proposed method. Specifically, we report the Equal False Positive Rate ($F_{FPR}$) \cite{ju2023improving}, Max Equalized Odds ($F_{MEO}$) \cite{wang2023aleatoric}, Demographic Parity ($F_{DP}$) \cite{wang2022understanding} and Overall Accuracy Equality ($F_{OAE}$) \cite{wang2023aleatoric}. The definition of those fairness metrics can be found in Appendix~\ref{appendix:fairness_metrics}.

\smallskip
\noindent
\textbf{Baseline Methods}.
We compare our method against the latest fairness method DAW-FDD \cite{ju2023improving} in deepfake detection. The comparison also includes `Ori' (a backbone with cross-entropy loss) and UCF \cite{Yan_2023_ICCV} (the latest disentanglement-based deepfake detector). Unless explicitly specified, all methods are employed on Xception \cite{chollet2017xception} backbone. 

\smallskip
\noindent
\textbf{Implementation Details}.
All experiments are based on the PyTorch and trained with NVIDIA RTX 3090Ti. For training, we fix the batch size 16, epochs 100, use SGD optimizer with learning rate $\beta=$\( 5 \times 10^{-4} \). For the overall loss, we set the $\lambda$ in Eq.~(\ref{eq:epsion_star}) as 1.0, the $\gamma$ (neighborhood size of perturbation in flattening loss) as 0.05, the $\rho_1$, $\rho_2$ in $L_{cls}$ as 0.1, 0.1, the $\rho_3$, $\rho_4$ in $\mathcal{L}_{dis}$ as 0.05 and 0.3, $b$ in $L_{con}$ as 3.0, and $\delta$ in  $M(\widehat{h}(d_i),D_i)$ as 2.89 based on the demographic sample distribution. The $\alpha$ and $\alpha'$ in  Eq.~(\ref{eq:fairness})  are tuned on the grid \{0,1,0.3,0.5,0.7,0.9\}. Following \cite{ju2023improving}, the final $\alpha$ and $\alpha'$ are determined based on a preset rule that allows up to a 5\% degradation of overall AUC in the validation set from the corresponding `Ori' method while minimizing the $F_{F\!P\!R}$ on Intersection group.

\begin{table}[]
\scalebox{0.70}{
\begin{tabular}{c|c|cccc|c}
\hline
\multirow{2}{*}{Testing Set} & \multirow{2}{*}{Method} & \multicolumn{4}{c|}{Fairness Metrics(\%)\textdownarrow} & \begin{tabular}[c]{@{}c@{}}Detection\\  Metric(\%)\textuparrow \end{tabular} \\ \cline{3-7} 
 &  & $F_{FPR}$ & $F_{MEO}$ & $F_{DP}$ & $F_{OAE}$ & AUC \\ \hline
\multirow{2}{*}{F2F \cite{thies2016face2face}} & DAW-FDD~\cite{ju2023improving} & 20.42 & 12.66 & 35.46 & 11.58 & 97.74 \\
 & \baseline{Ours} & \baseline{\textbf{17.42}} & \baseline{\textbf{10.00}} & \baseline{\textbf{33.20}} & \baseline{\textbf{9.56}} & \baseline{\textbf{98.65}} \\ \hline
\multirow{2}{*}{FS \cite{FaceSwap2018}} & DAW-FDD~\cite{ju2023improving} & 32.96 & 14.52 & 21.39 & \textbf{3.95} & 98.62 \\
 & \baseline{Ours} & \baseline{\textbf{26.32}} & \baseline{\textbf{9.97}} &\baseline{\textbf{19.30}} & \baseline{6.70} & \baseline{\textbf{99.23}} \\ \hline
\multirow{2}{*}{NT \cite{thies2019deferred}} & DAW-FDD~\cite{ju2023improving} & \textbf{23.64} & 20.83 & 20.50 & 17.36 & 94.99 \\
 & \baseline{Ours} & \baseline{23.98} & \baseline{\textbf{16.83}} & \baseline{\textbf{16.03}} & \baseline{\textbf{13.61}} & \baseline{\textbf{96.35}} \\ \hline
\multirow{2}{*}{DF \cite{DeepFakes2017}} & DAW-FDD~\cite{ju2023improving} & 20.41 & 12.66 & 9.99 & 6.16 & 98.20 \\
 & \baseline{Ours} & \baseline{\textbf{17.42}} & \baseline{\textbf{9.02}} & \baseline{\textbf{9.43}} & \baseline{\textbf{5.86}} & \baseline{\textbf{99.05}} \\ \hline
\multirow{2}{*}{FST \cite{li2019faceshifter}} & DAW-FDD~\cite{ju2023improving} & 25.36 & 10.05 & 10.34 & 8.79 & 98.02 \\
 & \baseline{Ours} & \baseline{\textbf{15.38}} &\baseline{\textbf{7.79}} & \baseline{\textbf{6.45}} & \baseline{\textbf{5.70}} & \baseline{\textbf{98.96}} \\ \hline
\end{tabular}
}
\vspace{-3mm}
\caption{Intra-domain evaluation on FF++. DAW-FDD and our method are trained on FF++, tested on its test sub-datasets separated by five forgeries, \textit{i.e.},\ F2F is the sub-dataset in FF++ test set generated by Face2Face~\cite{thies2016face2face}. The best results are shown in \textbf{Bold}.}
\label{tab:in_domain_different_forgery}
\vspace{-5mm}
\end{table}

\begin{table*}[h]
\centering
\scalebox{0.66}{
\begin{tabular}{c|c|ccccc|ccccc|ccccc}
\hline
\multirow{3}{*}{Dataset} & \multirow{3}{*}{Method} & \multicolumn{5}{c|}{Xception~\cite{chollet2017xception}} & \multicolumn{5}{c|}{ResNet-50~\cite{he2016deep}} & \multicolumn{5}{c}{EfficientNet-B3~\cite{tan2019efficientnet}} \\ \cline{3-17} 
 &  & \multicolumn{4}{c|}{Fairness Metrics(\%)\textdownarrow} & \begin{tabular}[c]{@{}c@{}}Detection \\ Metric(\%)\textuparrow \end{tabular} & \multicolumn{4}{c|}{Fairness Metrics(\%)\textdownarrow} & \begin{tabular}[c]{@{}c@{}}Detection \\ Metric(\%)\textuparrow \end{tabular} & \multicolumn{4}{c|}{Fairness Metrics(\%)\textdownarrow} & \begin{tabular}[c]{@{}c@{}}Detection \\ Metric(\%)\textuparrow \end{tabular} \\ \cline{3-17} 
 &  & $F_{FPR}$ & $F_{MEO}$ & $F_{DP}$ & \multicolumn{1}{c|}{$F_{OAE}$} & AUC & $F_{FPR}$ & $F_{MEO}$ & $F_{DP}$ & \multicolumn{1}{c|}{$F_{OAE}$} & AUC &$F_{FPR}$ & $F_{MEO}$ & $F_{DP}$ & \multicolumn{1}{c|}{$F_{OAE}$} & AUC \\ \hline
\multirow{4}{*}{FF++} & Ori~\cite{rossler2019faceforensics++} & 31.31 & 17.69 & 11.12 & \multicolumn{1}{c|}{10.08} & 92.77 & 34.69 & 17.29 & 9.83 & \multicolumn{1}{c|}{8.85} & 94.83 & 18.78 & 33.21 & 31.36 & \multicolumn{1}{c|}{26.01} & 93.55 \\
 & DAW-FDD~\cite{ju2023improving} & 14.06 & 10.55 & 10.97 & \multicolumn{1}{c|}{8.72} & 97.46 & 30.36 & 9.74 & 8.89 & \multicolumn{1}{c|}{7.42} & 93.23 & 23.33 & 26.15 & 24.74 & \multicolumn{1}{c|}{21.23} &94.92  \\
 & UCF~\cite{Yan_2023_ICCV} & 21.52 & 13.06 & 15.06 & \multicolumn{1}{c|}{10.58} & 97.10 & 35.13 & 10.87 & 10.81 & \multicolumn{1}{c|}{8.05} &95.92  & 20.92 & 33.08 & 30.01 & \multicolumn{1}{c|}{24.56} & 94.21 \\ \cline{2-17} 
 & \baseline{Ours} & \baseline{\textbf{10.63}}&\baseline{\textbf{8.15}} & \baseline{\textbf{10.41}} & \multicolumn{1}{c|}{\baseline{\textbf{7.60}}} & \baseline{\textbf{98.28}} & \baseline{\textbf{22.70}} & \baseline{\textbf{9.28}} & \baseline{\textbf{8.72}} & \multicolumn{1}{c|}{\baseline{\textbf{5.74}}} & \baseline{\textbf{97.72}} & \baseline{\textbf{11.19}} & \baseline{\textbf{20.61}} & \baseline{\textbf{18.40}} & \multicolumn{1}{c|}{\baseline{\textbf{16.18}}} & \baseline{\textbf{95.39}} \\ \hline
\multirow{4}{*}{DFDC} & Ori~\cite{rossler2019faceforensics++} & 52.77 & 37.78 & 13.87 & \multicolumn{1}{c|}{30.30} & 56.72 & 45.84 & 28.89 & 16.67 & \multicolumn{1}{c|}{26.25} & 58.08 & 62.38 & 37.56 & 22.44  & \multicolumn{1}{c|}{25.93} & 57.81 \\
 & DAW-FDD~\cite{ju2023improving} & 45.14 & 35.77 & 18.59 & \multicolumn{1}{c|}{14.07} & 59.96 & 44.07 & 34.14 & 18.72 & \multicolumn{1}{c|}{24.58} & 60.11 & 50.73 & 43.79 & 18.31 & \multicolumn{1}{c|}{29.57} & 58.29 \\
 & UCF~\cite{Yan_2023_ICCV} & 53.07 & 44.44 & 15.70 & \multicolumn{1}{c|}{23.22} & 60.03 & 43.39 & 35.62 & 15.86 &  \multicolumn{1}{c|}{19.15} & \textbf{61.06} & 42.79 & 40.54 & 19.35 & \multicolumn{1}{c|}{21.13} & 58.85 \\ \cline{2-17} 
 & \baseline{Ours} & \baseline{\textbf{40.73}} & \baseline{\textbf{34.48}} & \baseline{\textbf{9.69}} & \multicolumn{1}{c|}{\baseline{\textbf{13.71}}} & \baseline{\textbf{61.47}} & \baseline{\textbf{37.17}} & \baseline{\textbf{27.78}} &\baseline {\textbf{10.94}} & \multicolumn{1}{c|}{\baseline{\textbf{18.52}}} & \baseline{59.76} & \baseline{\textbf{22.89}} & \baseline{\textbf{33.78}} & \baseline{\textbf{12.35}} & \multicolumn{1}{c|}{\baseline{\textbf{16.73}}} & \baseline{\textbf{60.67}} \\ \hline
\multirow{4}{*}{Celeb-DF} & Ori~\cite{rossler2019faceforensics++} & 27.55 & 25.65 & 17.74 & \multicolumn{1}{c|}{58.44} & 62.66 & 24.94 & 22.32 & 19.47 & \multicolumn{1}{c|}{48.62} & 70.64 & 30.86 & 27.47 & 19.15 & \multicolumn{1}{c|}{59.32} & 62.36 \\
 & DAW-FDD~\cite{ju2023improving} & 22.31 & 20.60 & \textbf{11.65} & \multicolumn{1}{c|}{49.71} & 69.55 & 26.82 & 21.93 & 20.80 & \multicolumn{1}{c|}{47.14} & 75.70 & 31.36 & 21.79 & 6.91 & \multicolumn{1}{c|}{\textbf{50.86}} & 70.14 \\
 & UCF~\cite{Yan_2023_ICCV} & 27.81 & 25.96 & 16.51 & \multicolumn{1}{c|}{48.63} & 71.73 & 32.17 & 28.28 & 19.38 & \multicolumn{1}{c|}{45.15} & 76.44 & 24.95 & 22.41 & 15.14 & \multicolumn{1}{c|}{58.48} & 72.65 \\ \cline{2-17} 
 & \baseline{Ours} & \baseline{\textbf{10.62}} & \baseline{\textbf{12.77}} &\baseline{15.04} & \multicolumn{1}{c|}{\baseline{\textbf{36.01}}} & \baseline{\textbf{74.42}} & \baseline{\textbf{11.55}} & \baseline{\textbf{17.01}} & \baseline{\textbf{17.21}} & \multicolumn{1}{c|}{\baseline{\textbf{29.58}}} & \baseline{\textbf{78.55}} & \baseline{\textbf{13.00}} & \baseline{\textbf{9.73}} & \baseline{\textbf{5.21}} & \multicolumn{1}{c|}{\baseline{55.74}} & \baseline{\textbf{75.32}} \\ \hline
\multirow{4}{*}{DFD} & Ori~\cite{rossler2019faceforensics++} & 35.14 & 28.52 & 15.31 & \multicolumn{1}{c|}{12.95} & 74.34 & 31.76 & 26.91 & 5.90 & \multicolumn{1}{c|}{28.48} & 76.02 & 39.37 & 38.57 & 20.01 & \multicolumn{1}{c|}{17.00} & 75.87 \\
 & DAW-FDD~\cite{ju2023improving} & 34.02 & 29.37 & 15.75 & \multicolumn{1}{c|}{11.31} & 71.42 & 33.05 & 24.24 & 7.12 & \multicolumn{1}{c|}{27.08} & 77.05 &32.72  & 28.74 & 17.12 & \multicolumn{1}{c|}{24.70} & 74.76 \\
 & UCF~\cite{Yan_2023_ICCV} & 42.66 & 33.41 & 20.24 & \multicolumn{1}{c|}{19.84} & 81.88 & 42.54 & 33.17 & 5.24 & \multicolumn{1}{c|}{30.98} & 78.97 & 36.59 & 27.32 & 25.83 & \multicolumn{1}{c|}{9.36} & 76.76 \\ \cline{2-17} 
 & \baseline{Ours} & \baseline{\textbf{26.08}} & \baseline{\textbf{21.37}} & \baseline{\textbf{11.65}} & \multicolumn{1}{c|}{\baseline{\textbf{8.37}}} & \baseline{\textbf{84.82}} & \baseline{\textbf{25.71}} & \baseline{\textbf{20.02}} & \baseline{\textbf{2.34}} & \multicolumn{1}{c|}{\baseline{\textbf{25.60}}} & \baseline{\textbf{79.67}} & \baseline{\textbf{29.34}} & \baseline{\textbf{24.52}} & \baseline{\textbf{11.46}} & \multicolumn{1}{c|}{\baseline{\textbf{5.11}}} & \baseline{\textbf{77.28}} \\ \hline
\end{tabular}
}
\vspace{-3mm}
\caption{Comparison with different methods in terms of improving fairness and detection generalization under both intra-domain (FF++) and cross-domain (DFDC, Celeb-DF, and DFD) scenarios. \textuparrow~means higher is better and \textdownarrow~means lower is better.}
\label{tab:comparison_with_different_method}
\end{table*}

\begin{table*}[h]
\centering
\scalebox{0.85}{
\begin{tabular}{c|ccccccccccccccc}
\hline
 & \multicolumn{7}{c|}{} & \multicolumn{8}{c}{Dataset} \\ \cline{9-16} 
 & \multicolumn{7}{c|}{\multirow{-2}{*}{Method}} & \multicolumn{2}{c|}{FF++} & \multicolumn{2}{c|}{DFDC} & \multicolumn{2}{c|}{Celeb-DF} & \multicolumn{2}{c}{DFD} \\ \cline{2-16} 
\multirow{-3}{*}{Effects} & \multicolumn{1}{c|}{Name} & Cls(CE) & Cls & Rec & Con & Ff & \multicolumn{1}{c|}{Lf} & $F_{FPR}$\textdownarrow & \multicolumn{1}{c|}{AUC\textuparrow} & $F_{FPR}$\textdownarrow & \multicolumn{1}{c|}{AUC\textuparrow} & $F_{FPR}$\textdownarrow & \multicolumn{1}{c|}{AUC\textuparrow} & $F_{FPR}$\textdownarrow & AUC\textuparrow \\ \hline
 & \multicolumn{1}{c|}{VariantA} &  & \checkmark &  &  & \checkmark & \multicolumn{1}{c|}{\checkmark} & 17.62 & \multicolumn{1}{c|}{98.06} & 43.24 & \multicolumn{1}{c|}{58.14} & 19.08  & \multicolumn{1}{c|}{68.38} & 27.81 & 81.98 \\
 & \multicolumn{1}{c|}{VariantB} &  & \checkmark & \checkmark &  & \checkmark & \multicolumn{1}{c|}{\checkmark} & 17.40 & \multicolumn{1}{c|}{98.24} & 41.44 & \multicolumn{1}{c|}{59.84} &  13.61& \multicolumn{1}{c|}{71.07} &26.52 & 82.08  \\
 & \multicolumn{1}{c|}{VariantC} &  & \checkmark &  &\checkmark  & \checkmark & \multicolumn{1}{c|}{\checkmark} &15.96 & \multicolumn{1}{c|}{97.93} & 44.01 & \multicolumn{1}{c|}{60.91} &  12.76& \multicolumn{1}{c|}{72.41} &26.36 & 84.19  \\
\multirow{-4}{*}{Dl} & \multicolumn{1}{c|}{VariantD} & \checkmark &  & \checkmark & \checkmark & \checkmark & \multicolumn{1}{c|}{\checkmark} & 16.58 & \multicolumn{1}{c|}{98.05} &  42.76 & \multicolumn{1}{c|}{60.16} & 14.04& \multicolumn{1}{c|}{ 74.14} & 29.57  & 84.66 \\ \hline
 & \multicolumn{1}{c|}{\baseline{Ours}} & \baseline{} & \baseline{\checkmark} & \baseline{\checkmark} & \baseline{\checkmark} & \baseline{\checkmark} & \multicolumn{1}{c|}{\baseline{\checkmark}} & \baseline{\textbf{10.63}} & \multicolumn{1}{c|}{\baseline{\textbf{98.28}}} & \baseline{\textbf{40.73}} & \multicolumn{1}{c|}{\baseline{\textbf{61.47}}} & \baseline{\textbf{10.62}} & \multicolumn{1}{c|}{\baseline{\textbf{74.42}}} & \baseline{\textbf{26.08}} & \baseline{\textbf{84.82}} \\ \hline
 & \multicolumn{1}{c|}{VariantE} &  & \checkmark & \checkmark & \checkmark & \checkmark & \multicolumn{1}{c|}{} & 13.93 & \multicolumn{1}{c|}{97.98} & 44.91 & \multicolumn{1}{c|}{60.10} &  18.56 & \multicolumn{1}{c|}{ 73.47} &31.34  & 81.44 \\
\multirow{-2}{*}{Ff\&Lf} & \multicolumn{1}{c|}{VariantF} &  & \checkmark & \checkmark & \checkmark &  & \multicolumn{1}{c|}{\checkmark} & 18.67 & \multicolumn{1}{c|}{98.04} & 41.17 & \multicolumn{1}{c|}{61.03} &  14.72 & \multicolumn{1}{c|}{ 71.43} & 30.08  & 82.46 \\ \hline
\end{tabular}
}
\vspace{-3mm}
\caption{Ablation study of the loss constraints in our disentanglement learning (Dl) module, and the effectiveness of our feature fusion (Ff) and loss flattening (Lf). `Cls', `Rec', and `Con' represent our classification loss, reconstruction loss, and contrastive loss, respectively. `Cls(CE)' means we replace our demographic distribution-aware margin loss with cross-entropy loss. All methods are only trained on FF++.}
\label{tab:ablation_method}
\vspace{-4mm}
\end{table*}

\subsection{Results}
\noindent
\textbf{Performance on Intra-domain sub-datasets}.
Intra-domain evaluation, conducted on individual forgery sub-dataset, assesses the model's proficiency in fitting the specific forgery sub-dataset. As illustrated in Table~\ref{tab:in_domain_different_forgery}, our disentanglement learning approach, which separates domain-specific forgery, guides the model not to overfit to a particular forgery domain. In general, our method enhances fairness and consistently achieves a higher AUC on each sub-dataset compared to DAW-FDD. This result suggests the effectiveness of eliminating domain-specific biases.

\smallskip
\noindent
\textbf{Performance of Fairness Generalization}.
Taking Xception backbone as an example, Table~\ref{tab:comparison_with_different_method} shows our method has superior fairness generalization ability compared to other methods, while simultaneously achieving the best detection results. Specifically, our method has an 8.63\% improvement in $F_{DP}$ on DFDC and enhances the $F_{FPR}$ by 11.69\% on Celeb-DF, 7.94\% on DFD compared with DAW-FDD~\cite{ju2023improving}. In addition, although DAW-FDD, as a fair detector, works well on FF++ compared to Ori, it underperforms Ori under certain cross-domain scenarios, with a notable 4.72\% decrease in $F_{DP}$ on DFDC and declines in $F_{MEO}$ and $F_{DP}$ on DFD.
UCF~\cite{Yan_2023_ICCV}, recognized as a state-of-the-art detector in improving detection generalization, surpasses  Ori and DAW-FDD in detection. However,  it fails to ensure fairness, as evidenced by its $F_{DP}$ being 3.94\% inferior to Ori's even in intra-domain testing, with all four fairness metrics on DFD performing worse than Ori. Overall, our method outperforms all compared methods across most fairness metrics, achieving the best in both fairness generalization and AUC.

\begin{figure*}[h]
    \centering
    \includegraphics[width=1\textwidth]{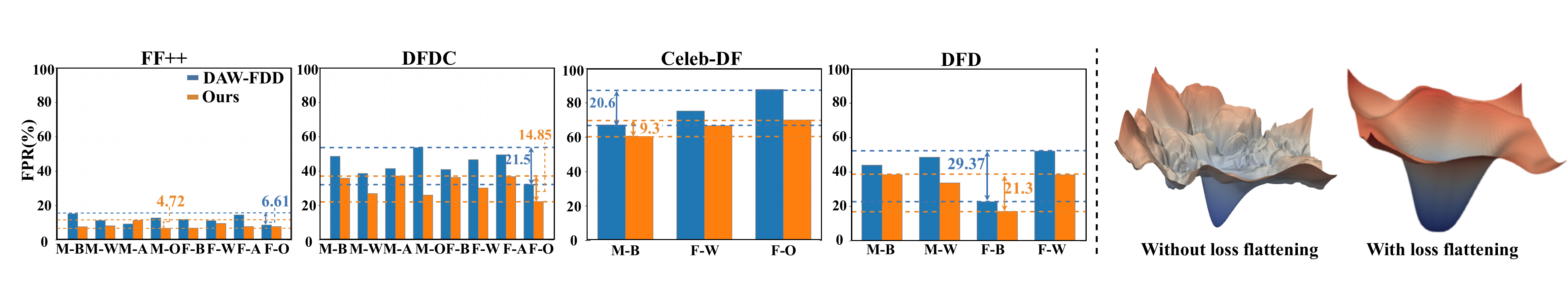}
    \vspace{-11mm}
    \caption{\small (\textbf{Left}) Comparison of FPR on Intersectional subgroups. Models are trained on FF++ and tested on FF++, DFDC, Celeb-DF, and DFD. The subgroups not represented in Celeb-DF and DFD are inapplicable. (\textbf{Right}) The loss landscape visualization of our proposed method with (right) and without (left) flattening the loss landscape.}
    \label{fig:fpr_loss}
\end{figure*}

\begin{figure*}[h]
    \centering
    \includegraphics[width=1\textwidth]{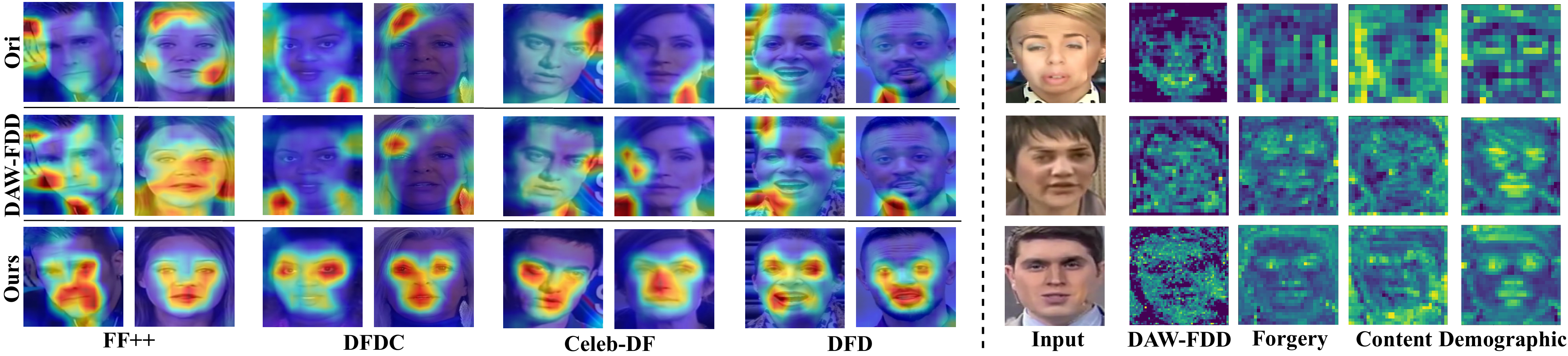}
    \vspace{-6mm}
    \caption{\small (\textbf{Left}) Grad-CAM visualization of Ori's (first row), DAW-FDD (second row), and ours (third row) on the intra-domain dataset (FF++), and cross-domain datasets (DFDC, Celeb-DF, and DFD). (\textbf{Right}) Visualization of the image (first column), DAW-FDD's features (second column), ours disentangled forgery (third column), content (fourth column), and demographic features (last column).}
    \label{fig:vis_one_row}
    \vspace{-4mm}
\end{figure*}

\smallskip
\noindent
\textbf{Fairness Generalization Performance of Different Backbones}.
To examine the fairness generalization capability of our proposed method concerning backbone selection, we substitute the Xception backbone with ResNet-50 \cite{he2016deep} and EfficientNet-B3 \cite{tan2019efficientnet}. The results in Table~\ref{tab:comparison_with_different_method} indicate that our method based on different backbones shows similar superior results. Such outcomes suggest that our proposed approach is not limited to backbone choice, but is effective and applicable to diverse backbone settings.

\subsection{Ablation Study} \label{subsec:ablation_study}

\smallskip
\noindent
\textbf{Effects of Components in Disentanglement Learning}. The results of VariantA/B/C/D in Table~\ref{tab:ablation_method} demonstrate the contribution of each loss constraint in our disentanglement learning (Dl) module. Without reconstructive loss and contrastive loss, VariantA shows relatively lower performance on both $F_{FPR}$ and AUC compared with other Variants and Ours. VariantB and VariantC underscore the value of our reconstructive loss (\emph{e.g.}, $F_{FPR}$ drops 5.47\% and the AUC increases 2.69\% on Celeb-DF) and contrastive loss (\emph{e.g.}, $F_{FPR}$ drops 6.32\% and the AUC increases 4.03\% on Celeb-DF), respectively. Comparing Ours with VariantD demonstrates the impact of our demographic distribution-aware margin loss. By replacing CE loss with the demographic distribution-aware margin loss, the $F_{FPR}$ reduces 5.95\% and the AUC improves 0.23\% on FF++. The similar tread is also observed on three other datasets.


\smallskip
\noindent
\textbf{Effects of Feature Fusion (Ff) and Loss Flattening (Lf)}. The results of VariantE/F in Table~\ref{tab:ablation_method} reveal the effects of our feature fusion (Ff) and loss flattening (Lf) methods. When comparing ours with VariantE (without Lf), the $F_{FPR}$ is enhanced by 7.94\% on Celeb-DF and 4.18\% on DFDC. While ours against VariantF (without Ff), the $F_{FPR}$ is improved by 4.10\% and 0.44\% on those two datasets. This indicates that Lf boosts the model's fairness generalization more than Ff. Overall, our method with both Ff and Lf yields the most substantial gains in fairness and AUC across all datasets.

\smallskip
\noindent
\textbf{Comparison on Intersectional Subgroups}.
We present detailed results of the False Positive Rate (FPR) on each subgroup across all datasets, as shown in Fig.~\ref{fig:fpr_loss} (left). The results clearly indicate that our approach significantly narrows the disparity between these subgroups, \emph{e.g.}, the maximum FPR gap of DAW-FDD on Celeb-DF is 20.6, while our method lowers the gap to 9.3. Overall, ours leads to a consistent and marked reduction in the FPR across all test datasets.


\subsection{Visualization}
\smallskip
\noindent
\textbf{Visualization of Loss Landscape}.
Fig.~\ref{fig:fpr_loss} (right) visually illustrates our method's loss landscape. Without the flattening process, the landscape is sharp with numerous peaks and valleys. Such sharpness may trap the model into different suboptimal minima, leading to inconsistent generalization. However, after flattening, the landscape becomes smoother, suggesting an easier optimization path, potentially leading to better training and generalization. This visualization underscores the significance of Joint Optimization in our method for enhancing fairness generalization.

\smallskip
\noindent
\textbf{Visualization of the Saliency Map}.
To more intuitively demonstrate the effectiveness of our method, we visualize the Grad-CAM~\cite{selvaraju2017gradcam} of Ori, DAW-FDD \cite{Yan_2023_ICCV}, and our method, respectively, as shown in Fig.~\ref{fig:vis_one_row} (left). Grad-CAM shows that the Ori without any constraints, is prone to overfitting to small local regions or focusing on content noise outside the facial region. DAW-FDD has the fair loss as a constraint that performs well in intra-domain. Once the data is unseen, it loses fair detection ability and its Grad-CAM shows similar results as Ori's. On the contrary, our method's activation region demonstrates a consistent model focus on facial salient features, irrespective of the dataset. 

\smallskip
\noindent
\textbf{Visualization of Features}.
The feature visualization in Fig.~\ref{fig:vis_one_row} (right) reveals key insights into the focus areas of DAW-FDD and our method. DAW-FDD's abstracted patterns and highlighted regions (second column) show a broad emphasis on facial features without specific targeting. In contrast, our disentangled features demonstrate distinct areas of focus: the forgery features (third column) and demographic features (last column) predominantly highlight facial areas, whereas the content features (fourth column) are oriented towards the background. This differentiation underscores the importance of integrating forgery and demographic features, and eliminating content features, to foster fairer learning.

\section{Conclusion}
While current methods for enhancing fairness in deepfake detection perform well within a specific domain, they struggle to maintain fairness when tested across different domains. Recognizing this limitation, we introduce an innovative framework designed to address the fairness generalization challenge in deepfake detection. By combining disentanglement learning and fair learning modules, our approach ensures both generalizability and fairness. Furthermore, we incorporate a loss flattening strategy to streamline the optimization process for these modules, resulting in robust fairness generalization. Experimental results on diverse deepfake datasets showcase the superior fairness maintenance capabilities of our method across various domains.

\smallskip
\noindent
\textbf{Limitation}.
One limitation of our method is its dependency on datasets including forged videos generated by multiple manipulation techniques. However, there exist few deepfake datasets that do not have such characteristics. 


\smallskip
\noindent
\textbf{Future Work}.
We aim to design a method that can preserve fairness not rely on multi-forged data, but can directly detect images generated by diffusion or GANs. In addition, we plan to enhance fairness across not just video datasets, but also in a multi-modal context.

{
    \small
    \bibliographystyle{ieeetr}
    \bibliography{main}
}

\newpage
\input{Appendix}

\end{document}

%% file: preamble.tex
\usepackage[dvipsnames]{xcolor}


%% file: Appendix.tex
\onecolumn
\appendix
\numberwithin{equation}{section}
\numberwithin{theorem}{section}
\numberwithin{figure}{section}
\numberwithin{table}{section}
\renewcommand{\thesection}{{\Alph{section}}}
\renewcommand{\thesubsection}{\Alph{section}.\arabic{subsection}}
\renewcommand{\thesubsubsection}{\Roman{section}.\arabic{subsection}.\arabic{subsubsection}}

\def\p{\mathbf{p}}
\def\v{\mathbf{v}}
\def\u{\mathbf{u}}

\begin{center}
\textbf{\Large Appendix for ``Preserving Fairness Generalization in Deepfake Detection"}
\end{center}

\section{Related Work}\label{appendix:related_works}
\textbf{Deepfake Detection}. Current deepfake detection methods can be categorized into three primary groups based on the features they employ. The first category hinges on identifying inconsistencies in the \textit{physical and physiological} characteristics of deepfakes. For example, inconsistent corneal specular highlights \cite{hu2021exposing}, the irregularity of pupil shapes \cite{guo2022eyes,guo2022open}, eye blinking patterns \cite{li2018ictu}, eye color difference \cite{matern2019exploiting}, facial landmark locations \cite{yang2019exposing}, etc. 
The second category concentrates on \textit{signal-level} artifacts introduced during the synthesis process, especially those from the frequency domain \cite{qian2020thinking}. 
These methods encompass various techniques, such as examining disparities in the frequency spectrum \cite{khayatkhoei2022spatial, dzanic2020fourier}, utilizing checkerboard artifacts introduced by the transposed convolutional operator \cite{frank2020leveraging, zhang2019detecting}.
However, the methods from the above two categories usually exhibit relatively low detection performance. Therefore, the largest portion of existing detection methods fall into the \textit{data-driven} category, including \cite{marra2019incremental, goebel2020detection, wang2020cnn, liu2020global, hulzebosch2020detecting, guo2022robust, pu2022learning}. These methods leverage various types of Deep Neural Networks (DNNs) trained on both authentic and deepfake videos to capture specific discernible artifacts. While these methods have achieved promising performance for the intra-domain evaluation, their performance sharply degrades during cross-domain testing.

\noindent\textbf{Generalization in Deepfake Detection}.  To address the generalization issue, disentanglement learning \cite{wang2022disentangled} is widely used to extract the forgery-related features while getting rid of forgery-irrelated features for detection. For example, Hu et al. \cite{hu2021improving} propose a disentanglement framework to automatically locate the forgery-related region for detection. Based on this framework, Zhang et al. \cite{zhang2020face} add auxiliary supervision to improve the generalization ability. To enhance the independence of disentangled features, Liang et al. \cite{liang2022exploring} propose a new framework by introducing content consistency constraints and global representation contrastive
constraints. Such framework is later extended  \cite{Yan_2023_ICCV} by exclusively utilizing common forgery features, which are extracted separately from forgery-related features for detection.

\noindent\textbf{Fairness in Deepfake Detection}. Recent studies have delved into fairness concerns within the domain of deepfake detection \cite{masood2022deepfakes}. Trinh et al. \cite{trinh2021examination} examined biases in existing deepfake datasets and detection models across protected subgroups. They found a large error rate difference among subgroups, consistent with similar observations in the study \cite{hazirbas2021towards}. Pu et al. \cite{pu2022fairness} assessed the reliability of the deepfake detection model MesoInception-4 on FF++ and revealed its overall unfairness toward both genders. 
A more comprehensive analysis of deepfake detection bias, encompassing both demographic and non-demographic attributes, was presented by Xu et al. \cite{xu2022comprehensive}. The authors significantly enriched five widely used deepfake detection datasets with diverse annotations to facilitate future research in this area. 
Furthermore, \cite{nadimpalli2022gbdf} highlighted substantial bias in both datasets and detection models. In an effort to mitigate performance bias across genders, they introduced a gender-balanced dataset. However, this approach yielded only modest improvements and required extensive data annotation efforts. 
More recently, Ju et al. \cite{ju2023improving} enhance fairness in testing scenarios within the same data domain, they do not maintain fairness when applied to cross-domain testing, which is the central focus of this paper.  

\section{Fairness Metrics}\label{appendix:fairness_metrics}
We assume a test set comprising indices \{1, \ldots, $n$\}. $Y_j$ and $\hat{Y}_j$ respectively represent the true and predicted labels of the sample $X_j$. Their values are binary, where 0 means real and 1 means fake. For all fairness metrics, a lower value means better performance.

\begin{align*}\small
    &{F_{F\!P\!R}} := \sum_{\mathcal{J}_j\in\mathcal{J}} \left| \frac{\sum_{j=1}^{n} \mathbb{I}_{[\hat{Y}_j = 1, D_j =  \mathcal{J}_j, Y_j = 0]}}{\sum_{j=1}^{n} \mathbb{I}_{[D_j = \mathcal{J}_j, Y_j = 0]}} - \frac{\sum_{j=1}^{n} \mathbb{I}_{[\hat{Y}_j = 1, Y_j = 0]}}{\sum_{j=1}^{n} \mathbb{I}_{[Y_j = 0]}} \right|,\\
    &F_{O\!A\!E} := \max_{\mathcal{J}_j\in\mathcal{J}} \left\{ \frac{\sum_{j=1}^{n} \mathbb{I}_{[\hat{Y}_j = Y_j, D_j = \mathcal{J}_j]}}{\sum_{j=1}^{n} \mathbb{I}_{[D_j = \mathcal{J}_j]}} \right. \quad \left. - \min_{{\mathcal{J}_j}' \in \mathcal{J}} \frac{\sum_{j=1}^{n} \mathbb{I}_{[\hat{Y}_j = Y_j, D_j = {\mathcal{J}_j}']}}{\sum_{j=1}^{n} \mathbb{I}_{[D_j = {\mathcal{J}_j}']}} \right\},\\
    &F_{DP} := \max_{k \in \{0,1\}} \left\{ \max_{J_j \in \mathcal{J}} \frac{\sum_{j=1}^{n} \mathbb{I}_{[\hat{Y}_j=k, D_j=J_j]}}{\sum_{j=1}^{n} \mathbb{I}_{[D_j=J_j]}} \right.\quad \left. - \min_{J_j' \in \mathcal{J}} \frac{\sum_{j=1}^{n} \mathbb{I}_{[\hat{Y}_j=k, D_j=J_j']}}{\sum_{j=1}^{n} \mathbb{I}_{[D_j=J_j']}} \right\}, \\
    &F_{M\!E\!O} := \max_{k, k' \in \{0,1\}} \left\{ \max_{J_j \in \mathcal{J}} \frac{\sum_{j=1}^{n} \mathbb{I}_{[\hat{Y}_j=k, Y_j=k', D_j=J_j]}}{\sum_{j=1}^{n} \mathbb{I}_{[D_j=J_j, Y_j=k]}} \right. \quad \left. - \min_{J_j' \in \mathcal{J}} \frac{\sum_{j=1}^{n} \mathbb{I}_{[\hat{Y}_j=k, Y_j=k', D_j=J_j']}}{\sum_{j=1}^{n} \mathbb{I}_{[D_j=J_j', Y_j=k]}} \right\}.
\end{align*}

Where $D$ is the demographic variable, $\mathcal{J}$ is the set of subgroups with each subgroup $\mathcal{J}_j\in\mathcal{J}$. $F_{FPR}$ meatures the disparity in False Positive Rate (FPR) across different groups compared to the overall population.  $F_{OAE}$ meatures the maximum ACC gap across all demographic groups. $F_{DP}$ measures the maximum difference in prediction rates across all demographic groups. And $F_{MEO}$ captures the largest disparity in prediction outcomes (either positive or negative) when comparing different demographic groups.

\section{The Network Details}\label{appendix:network_details}
\begin{figure}[h]
  \centering
  \includegraphics[width=1\linewidth]{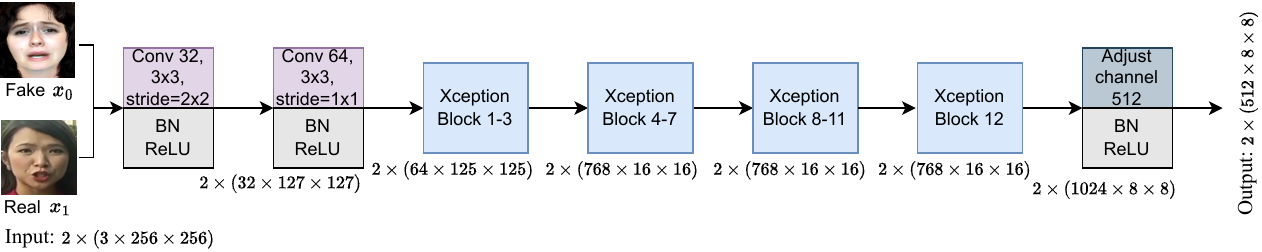}
  \vspace{-5mm}
  \caption{The architecture details of the encoder in our proposed method.}
  \label{fig:encoder}
\end{figure}

\textbf{Encoder}.
The architecture details of the encoder in our proposed method are presented in Fig.~\ref{fig:encoder}. An image pair, comprising one fake and one real image, serves as the input, which is subsequently processed by an encoder built upon the Xception~\cite{chollet2017xception} backbone.

\begin{figure}[h]
  \centering
  \includegraphics[width=1\linewidth]{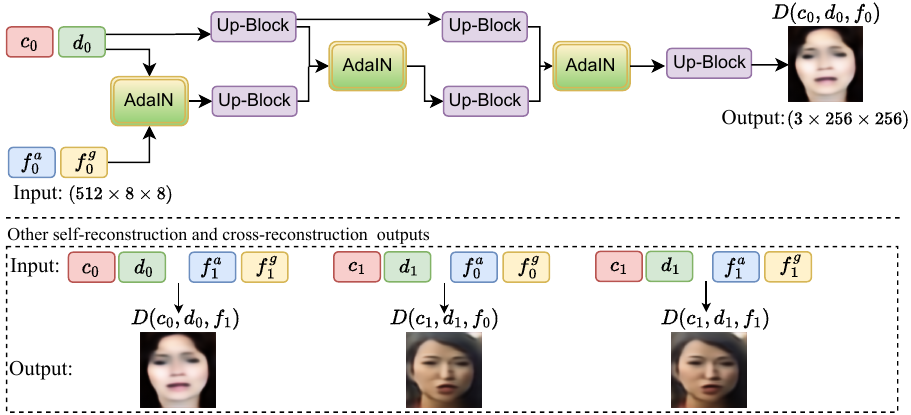}
  \vspace{-5mm}
  \caption{The architecture details of the decoder in our proposed method.}
  \label{fig:decoder}
\end{figure}

\smallskip
\noindent
\textbf{Decoder}.
We further present the architecture details of the decoder in Fig.~\ref{fig:decoder}, which reconstructs images in our proposed method to preserve the integrity of the extracted features. The demographic features $d_0$ and the content features $C_0$ are extracted from encoder, while $f_0^a$ and $f_0^g$ represent the domain-specific features and domain-agnostic features, respectively. The decoder reconstructs an image by utilizing those features separated by our disentanglement learning module as input, and passes through a series of upsampling and convolutional layers (Up-Block). AdaIN~\cite{huang2017arbitrary} is applied here for improving reconstructing and decoding. We present more visualizations of reconstruction images in different training epochs. We observe that, as the training progresses, the model learns to capture more detail features (\emph{e.g.}, facial characteristics). This further validates our decoder successfully preserves the completeness of the extracted features.

\begin{figure}[h]
  \centering
  \includegraphics[width=0.5\linewidth]{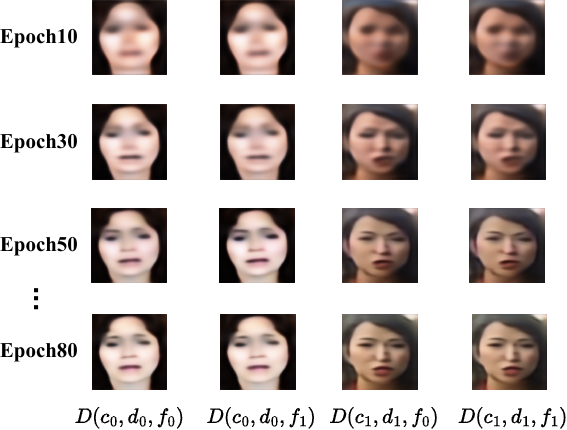}
  \vspace{-2mm}
  \caption{Visualization of the reconstruction images during the training process.}
  \label{fig:rec_process}
\end{figure}

\section{End-to-end Training Algorithm}\label{appendix:algorithm_details}
Below is the pseudocode of our joint optimization, which integrates a loss flattening strategy based on sharpness-aware minimization~\cite{foret2020sharpness}, and is implemented throughout the end-to-end training process.

\begin{algorithm}[h]\footnotesize
    \caption{Joint Optimization}\label{appendix:Joint Optimization}
    \SetAlgoLined
    \KwIn{A training dataset $\mathcal{S}$ with demographic variable {$D$}, a set of subgroups $\mathcal{J}$,  $\alpha$, $\alpha'$, max\_iterations, num\_batch, learning rate $\beta$}
    \KwOut{A deepfake detection model with fairness generalizability} 
    
    \textbf{Initialization:} $\theta_0$, $l=0$   

    
    \For{$e=1$ to \emph{max\_iterations}}{
    \For{$b=1$ to \emph{num\_batch}}{
    { \mbox{Sample a mini-batch $\mathcal{S}_b$ from $\mathcal{S}$}}
    
    Compute sample loss of $(C(h(I_i), Y_i))$, $\forall (I_i,Y_i)\in \mathcal{S}_b$
    
    For each $j\in\{1,...,|\mathcal{J}|\}$, set $\eta_j^*$ to be the value of $\eta_j$ that minimizes $L_j$ as given in (\ref{eq:fairness2}). This minimization is solved using binary search.
    
    Set $L_j(\theta)\leftarrow L_j(\theta,\eta_j^*)$ using (\ref{eq:fairness2}), $\forall j$
    
    Using binary search to find $\eta$ that minimizes (\ref{eq:fairness1})

    Compute $\epsilon^*$ based on Eq. (\ref{eq:epsion_star})

    Compute gradient approximation for (\ref{eq:sharpness})
    
    Update $\theta$: $\theta_{l+1}\leftarrow\theta_{l}-\beta \nabla_\theta \mathcal{L}\big|_{\theta_l+\epsilon^*}$

    $l\leftarrow l+1$
    
    }
    }
    \Return{$\theta_{l}$}
\end{algorithm}

\section{Additional Experimental Settings}\label{appendix:additional_settings}
We show the total number of train, validation and test samples of each dataset and the attributes included in our experiment in Table~\ref{tab:dataset_details}. We only use FF++ for training and validation.

\begin{table}[h]
\centering
\begin{tabular}{c|ccc|c}
\hline
\multirow{2}{*}{Dataset} & \multicolumn{3}{c|}{Samples} & \multirow{2}{*}{Intersection Sensitive Attributes} \\ \cline{2-4}
 & Train & Validation & Test &  \\ \hline
FF++ & 76,139 & 25,386 & 25,401 & M-A, M-B, M-W, M-O, F-A, F-B, F-W, F-O \\ \hline
DFD & - & - & 9,385 & M-B, M-W, M-O, F-B, F-W, F-O \\ \hline
DFDC & - & - & 22,857 & M-A, M-B, M-W, M-O, F-A, F-B, F-W, F-O \\ \hline
Celeb-DF & - & - & 28,458 & M-B, M-W, M-O, F-B, F-W, F-O \\ \hline
\end{tabular}
\caption{Test sample number and Intersection attributes in each dataset. `-' means not used.}
\label{tab:dataset_details}
\end{table}

\section{Additional Experimental Results}\label{appendix:additional_results}
\textbf{Stability Evaluation}.
The stability comparison of DAW-FDD with ours over 5 random runs is shown in Table~\ref{tab:stability_evaluation}. Our method shows superior fairness and detection mean score out of 5 random runs compared to DAW-FDD. This suggests that our approach has a robust and formidable capacity to improve fairness.

\begin{figure}[h]
  \centering
  \includegraphics[width=0.8\linewidth]{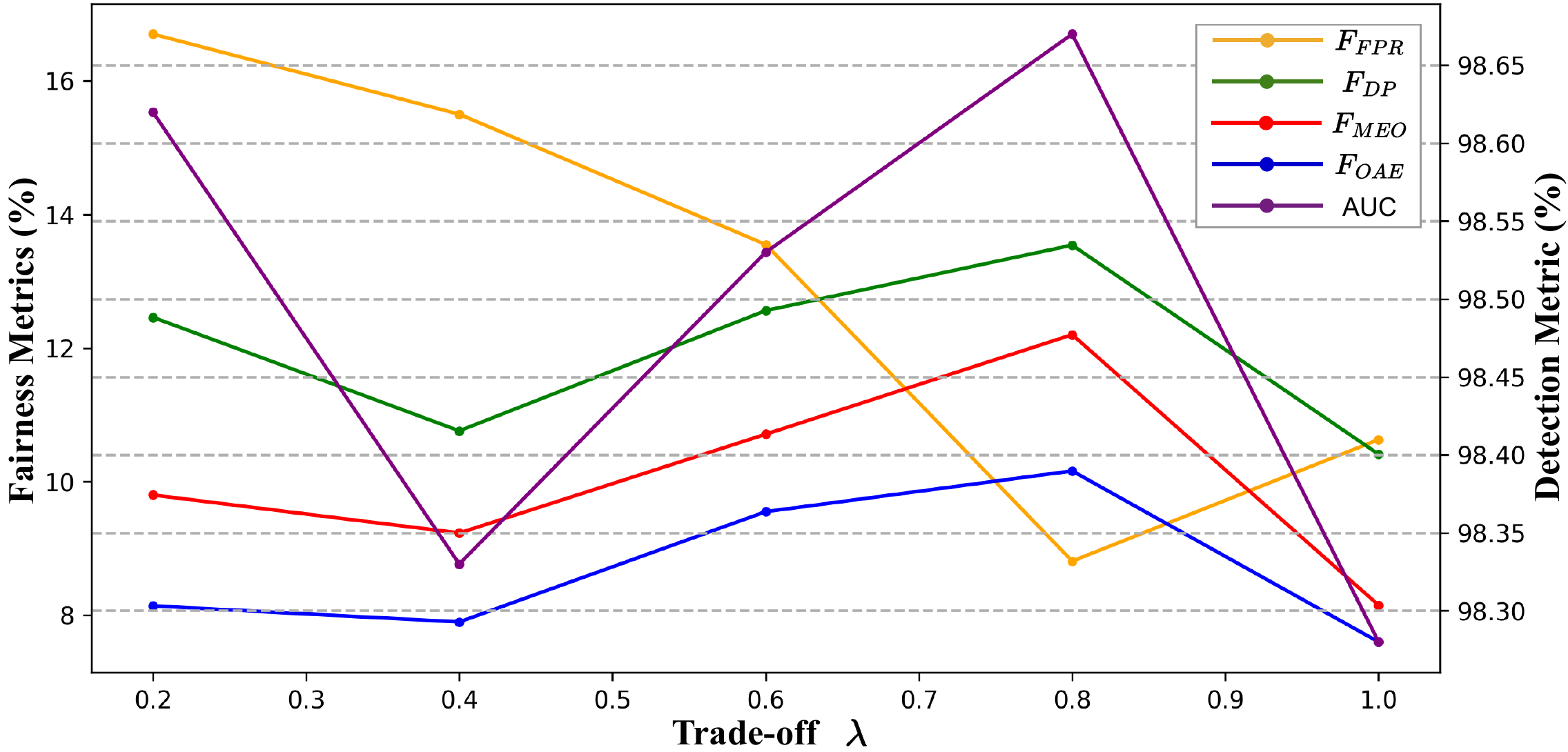}
  \vspace{-2mm}
  \caption{Sensitivity analysis of parameter $\lambda$ on the trade-off between fairness and detection accuracy on FF++.}
  \label{fig:sensitivity}
  \vspace{-2mm}
\end{figure}

\begin{figure}[h]
  \centering
  \includegraphics[width=1\linewidth]{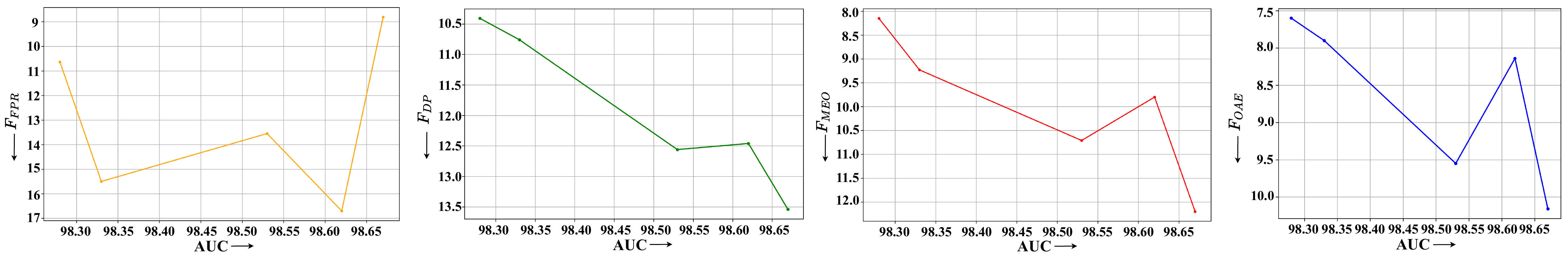}
  \vspace{-5mm}
  \caption{Trends in Fairness Metrics vs. AUC Score. From left to right, the graphs show how $F_{FPR}$, $F_{DP}$, $F_{MEO}$, and $F_{OAE}$ change with AUC, illustrating the trade-off between accuracy and fairness.}
  \label{fig:fairness_auc}
\end{figure}

\smallskip
\noindent
\textbf{Effect of Trade-off $\lambda$}.
To validate the effect of the trade-off hyperparameter in Eq.~\ref{eq:epsion_star}, we conduct sensitivity analysis on FF++ dataset. Fig.~\ref{fig:sensitivity} shows the fairness metrics and detection metric AUC to different $\lambda$ values. Experiment results demonstrate that the model attains optimal fairness performance when $\lambda$ is configured to 1.0 and also keeps fair AUC score. Notably, the analysis uncovers a trade-off between fairness and AUC score: as $\lambda$ ranges from 0.4 to 0.8, there is an enhancement in AUC while the fairness ($F_{DP}$, $F_{MEO}$, and $F_{OAE}$) becomes worse. However, when $\lambda$ changes from 0.8 to 1.0, we can see the opposite effect: AUC decreases while the fairness improves. Specifically, the behavior of $F_{FPR}$ diverges from that of the other fairness metrics. This is because a higher AUC typically reflects an optimal balance between maximizing the TPR and minimizing the FPR. As a result, at a $\lambda$ of 0.8, a lower $F_{FPR}$ is accompanied by a higher AUC. To more clearly show the relationship between each fairness metric and AUC, we present these dynamics separately in Fig.~\ref{fig:fairness_auc}, which illustrates the trend where gains in AUC correspond to diminished fairness.
\begin{table}[]
\scalebox{0.55}{
\begin{tabular}{c|ccccc|ccccc|ccccc|ccccc}
\hline
\multirow{3}{*}{Method} & \multicolumn{5}{c|}{FF++} & \multicolumn{5}{c|}{DFDC} & \multicolumn{5}{c|}{Celeb-DF} & \multicolumn{5}{c}{DFD} \\ \cline{2-21} 
 & \multicolumn{4}{c|}{Fairness Metrics(\%)\textdownarrow} & \begin{tabular}[c]{@{}c@{}}Detection \\ Metric(\%)\textuparrow\end{tabular} & \multicolumn{4}{c|}{Fairness Metrics(\%)\textdownarrow} & \begin{tabular}[c]{@{}c@{}}Detection\\  Metric(\%)\textuparrow\end{tabular} & \multicolumn{4}{c|}{Fairness Metrics(\%)\textdownarrow} & \begin{tabular}[c]{@{}c@{}}Detection\\  Metric(\%)\textuparrow\end{tabular} & \multicolumn{4}{c|}{Fairness Metrics(\%)\textdownarrow} & \begin{tabular}[c]{@{}c@{}}Detection\\  Metric(\%)\textuparrow\end{tabular} \\ \cline{2-21} 
 & $F_{FPR}$ & $F_{MEO}$ & $F_{DP}$ & \multicolumn{1}{c|}{$F_{OAE}$} & AUC & $F_{FPR}$ & $F_{MEO}$ & $F_{DP}$ & \multicolumn{1}{c|}{$F_{OAE}$} & AUC & $F_{FPR}$ & $F_{MEO}$ & $F_{DP}$ & \multicolumn{1}{c|}{$F_{OAE}$} & AUC & $F_{FPR}$ & $F_{MEO}$ & $F_{DP}$ & \multicolumn{1}{c|}{$F_{OAE}$} & AUC \\ \hline
\multirow{2}{*}{DAW-FDD} & 15.81  & 11.19 & 12.57 & \multicolumn{1}{c|}{9.66} & 97.54 & 44.97 & 35.07 & 16.19 & \multicolumn{1}{c|}{18.59} & 60.28 & 21.32 & 19.96 & 16.17 & \multicolumn{1}{c|}{49.44} & 69.97 & 34.69  & 29.36 & 18.59 & \multicolumn{1}{c|}{12.05} & 73.54 \\
 & (1.62)  & (2.48) & (2.15) & \multicolumn{1}{c|}{(2.11)} & (0.23) & (1.62) & (2.23) & (2.03) & \multicolumn{1}{c|}{(3.24)} & (1.11) & (4.63) & (5.34) & (7.01) & \multicolumn{1}{c|}{(8.43)} & (0.84) & (1.75) & (1.77) & (2.64) & \multicolumn{1}{c|}{(1.38)} & (2.45) \\ \hline
\multirow{2}{*}{Ours} & \textbf{11.70} & \textbf{10.40} & \textbf{11.93} & \multicolumn{1}{c|}{\textbf{8.73}} & \textbf{98.17} & \textbf{39.22} & \textbf{35.03} & \textbf{10.10} & \multicolumn{1}{c|}{\textbf{17.10}} & \textbf{61.84} & \textbf{10.93} & \textbf{12.58} & \textbf{13.52} & \multicolumn{1}{c|}{\textbf{34.05}} & \textbf{75.23} & \textbf{27.14} & \textbf{22.86} & \textbf{17.58} & \multicolumn{1}{c|}{\textbf{8.38}} & \textbf{82.79} \\
 & (1.89) & (1.96) & (1.46) & \multicolumn{1}{c|}{(1.38)} & (0.28) & (4.04) & (1.83) & (0.92) & \multicolumn{1}{c|}{(2.37)} & (0.66) & (4.79) & (2.56) & (4.12) & \multicolumn{1}{c|}{(7.37)} & (1.81) & (0.94) & (1.52) & (4.36) & \multicolumn{1}{c|}{(0.89)} & (2.50) \\ \hline
\end{tabular}
}
\caption{Detection mean and standard deviation (in parentheses) on intra-domain and cross-domain testing sets across 5 experimental repeats. Each method is trained only on FF++.}
\label{tab:stability_evaluation}
\end{table}

\begin{figure}[h]
  \centering
  \includegraphics[width=0.5\linewidth]{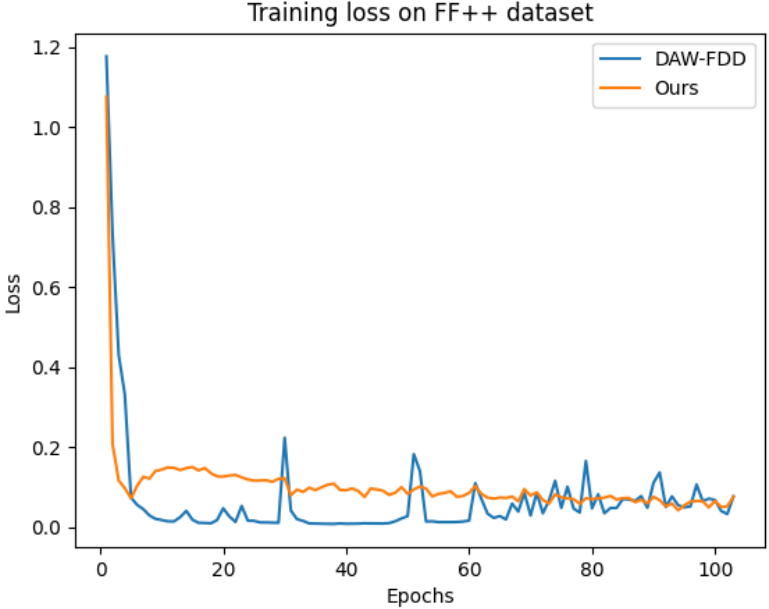}
  \vspace{-3mm}
  \caption{Training loss convergence.}
  \label{fig:loss_convergence}
\end{figure}

\smallskip
\noindent
\textbf{Comparison of the Loss Convergence}.
In Fig.~\ref{fig:loss_convergence}, we present a comparison of training loss convergence between our method and DAW-FDD, both utilizing Xception as the backbone on the FF++ dataset. It is evident that while DAW-FDD exhibits fluctuating convergence, our method demonstrates a more stable and consistent reduction in training loss. This stability indicates potential advantages in the robustness and reliability of our approach during the training process.

\begin{figure*}[t]
    \centering
    \includegraphics[width=1\textwidth]{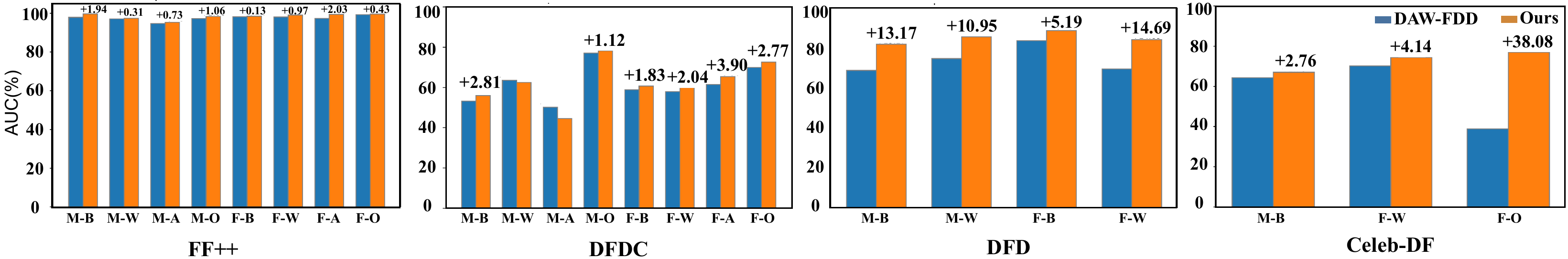}
    \vspace{-5mm}
    \caption{\small AUC comparison of DAW-FDD and Ours on the Intersectional subgroups. The subgroups not represented in DFD and Celeb-DF are inapplicable. }
    \label{fig:auc_on_inter}
\end{figure*}

\smallskip
\noindent
\textbf{Comparison of AUC on Intersectional Subgroups}.
We further show the AUC comparison results on FF++, DFDC, DFD, and Celeb-DF datasets with detailed performance in subgroups in Fig.~\ref{fig:auc_on_inter}. Our method evidently improves the AUC of each subgroup and narrows the disparity between subgroups. Notably, in DFD and Celeb-DF, the AUC difference between subgroups is much lower than DAW-FDD's.

\begin{figure}[h]
  \centering
  \includegraphics[width=0.5\linewidth]{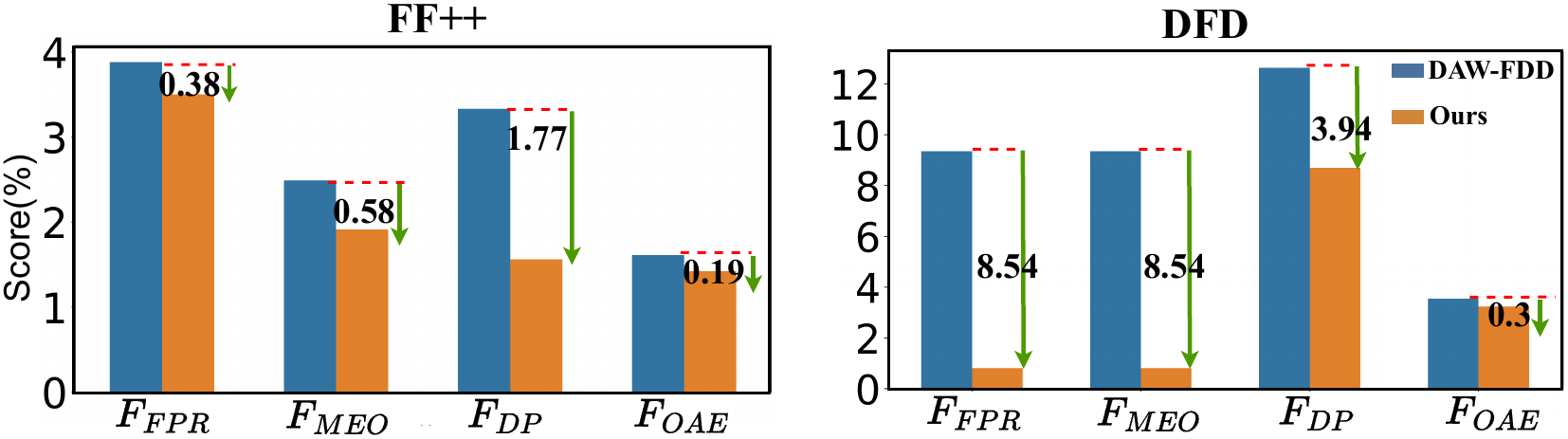}
  \vspace{-3mm}
  \caption{Comparison of fairness performance on Race subgroup (cross-domain and cross-subgroup). Models are trained on FF++ using Intersection attribute, tested on Celeb-DF and DFD under Race subgroup. }
  \label{fig:performance_on_race}
\end{figure}

\smallskip
\noindent
\textbf{Comparison on Cross-demographic Subgroup}.
DAW-FDD and our model are trained on FF++ with Intersection demographic information, tested on Celeb-DF and DFD, we report the fairness performance on the Race subgroup. The results shown in Fig.~\ref{fig:performance_on_race} clearly demonstrate that our method exhibits substantial improvements on $F_{FPR}$, $F_{MEO}$,  and $F_{OAE}$ fairness metrics, particularly noticeable on the $F_{FPR}$ and $F_{MEO}$ in DFD. This suggests that our approach can maintain fairness generalization ability among different demographic subgroups.

\begin{figure}[h]
  \centering
  \includegraphics[width=1\linewidth]{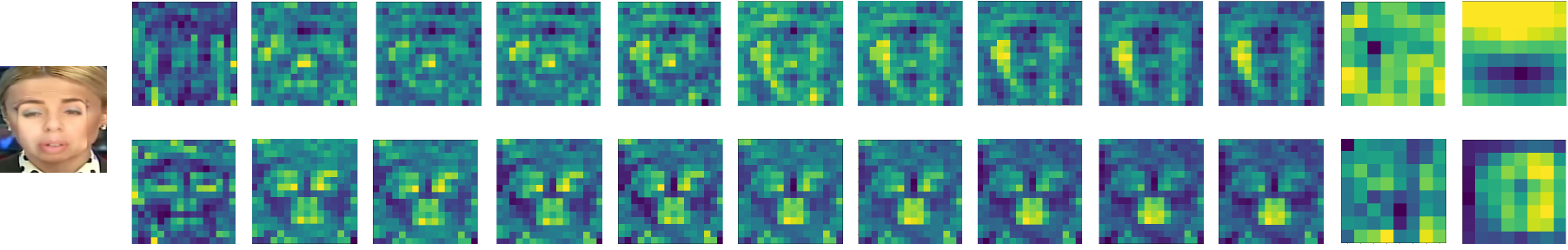}
  \vspace{-5mm}
  \caption{More visualization of our disentangled forgery features (first row) and demographic features (second row) from our method on FF++.}
  \label{fig:more_vis}
\end{figure}

\begin{figure}[h]
  \centering
  \includegraphics[width=0.4\linewidth]{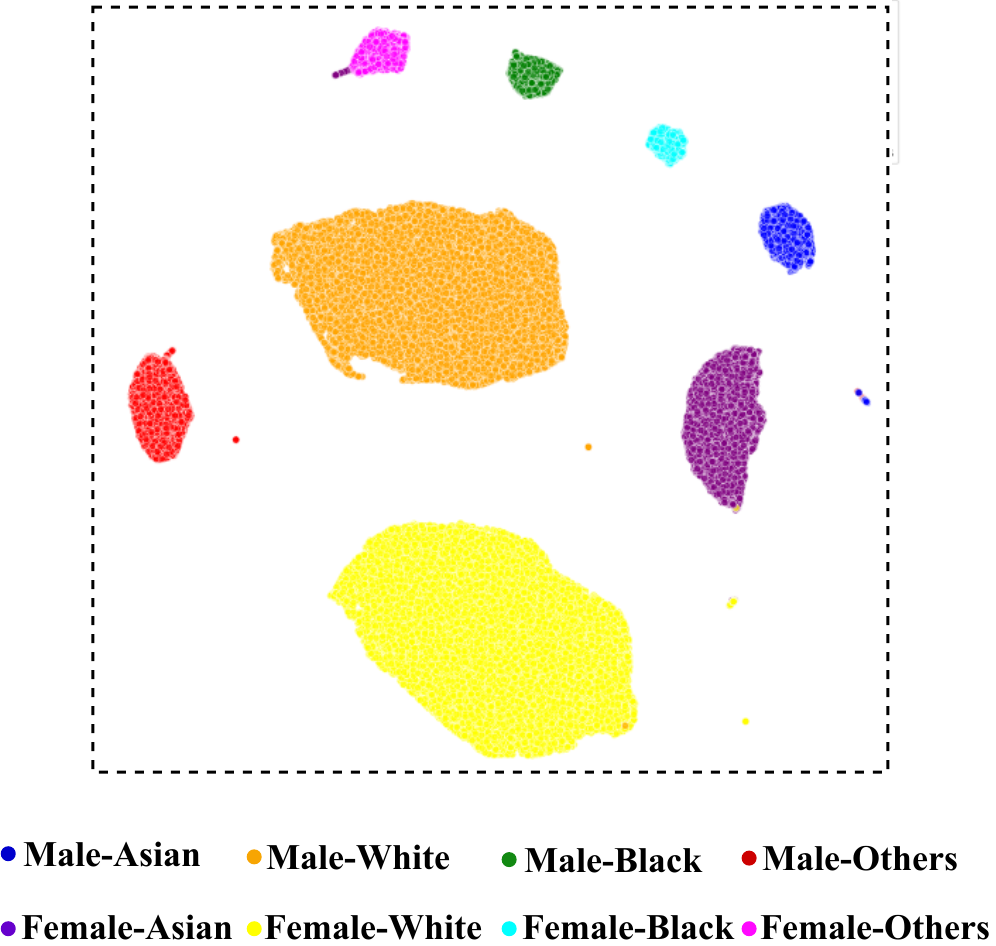}
  \vspace{-2mm}
  \caption{The UMAP~\cite{mcinnes2018umap} visualization of demographic features extracted from our method on FF++.}
  \label{fig:umap_demographic}
\end{figure}

\smallskip
\noindent
\textbf{Visualization}.
\textbf{1)} Detailed feature visualization of our disentangled forgery features and demographic features are presented in Fig.~\ref{fig:more_vis}. From left to right, the visualization demonstrates how our network builds up its understanding from original image. \textbf{2)} In addition, we show the UMAP~\cite{mcinnes2018umap} visualization of demographic features extracted from our method on FF++ in Fig.~\ref{fig:umap_demographic}. In the visualization, images with different intersectional demographic attributes locate separately in the latent space, which reveals that our model's capability to distinguish and disentangle features from different demographic backgrounds effectively. The result also aligns with demographic feature visualization in Fig.~\ref{fig:more_vis}, that our model actually captures demographic features for fair learning. The UMAP result further shows that the majority of subgroups in FF++ are Male-White and Female-White, the bias in the dataset makes it challenging for fair detection, suggesting the necessity of the demographic distribution-aware margin loss~\cite{cao2019learning} we apply in our method for improving generalization for minority subgroups.

%% file: main.bbl
\begin{thebibliography}{10}

\bibitem{vahdat2020nvae}
A.~Vahdat and J.~Kautz, ``Nvae: A deep hierarchical variational autoencoder,'' {\em Advances in neural information processing systems}, vol.~33, pp.~19667--19679, 2020.

\bibitem{karras2020analyzing}
T.~Karras, S.~Laine, M.~Aittala, J.~Hellsten, J.~Lehtinen, and T.~Aila, ``Analyzing and improving the image quality of stylegan,'' in {\em Proceedings of the IEEE/CVF conference on computer vision and pattern recognition}, pp.~8110--8119, 2020.

\bibitem{dhariwal2021diffusion}
P.~Dhariwal and A.~Nichol, ``Diffusion models beat gans on image synthesis,'' {\em Advances in neural information processing systems}, vol.~34, pp.~8780--8794, 2021.

\bibitem{wang2022gan}
X.~Wang, H.~Guo, S.~Hu, M.-C. Chang, and S.~Lyu, ``Gan-generated faces detection: A survey and new perspectives,'' {\em ECAI}, 2023.

\bibitem{masood2023deepfakes}
M.~Masood, M.~Nawaz, K.~M. Malik, A.~Javed, A.~Irtaza, and H.~Malik, ``Deepfakes generation and detection: State-of-the-art, open challenges, countermeasures, and way forward,'' {\em Applied intelligence}, vol.~53, no.~4, pp.~3974--4026, 2023.

\bibitem{ju2023improving}
Y.~Ju, S.~Hu, S.~Jia, G.~H. Chen, and S.~Lyu, ``Improving fairness in deepfake detection,'' in {\em Proceedings of the IEEE/CVF Winter Conference on Applications of Computer Vision}, pp.~4655--4665, 2024.

\bibitem{marra2019incremental}
F.~Marra, C.~Saltori, G.~Boato, and L.~Verdoliva, ``Incremental learning for the detection and classification of gan-generated images,'' in {\em 2019 IEEE international workshop on information forensics and security (WIFS)}, pp.~1--6, IEEE, 2019.

\bibitem{goebel2020detection}
M.~Goebel, L.~Nataraj, T.~Nanjundaswamy, T.~M. Mohammed, S.~Chandrasekaran, and B.~Manjunath, ``Detection, attribution and localization of gan generated images,'' {\em arXiv preprint arXiv:2007.10466}, 2020.

\bibitem{wang2020cnn}
S.-Y. Wang, O.~Wang, R.~Zhang, A.~Owens, and A.~A. Efros, ``Cnn-generated images are surprisingly easy to spot... for now,'' in {\em Proceedings of the IEEE/CVF conference on computer vision and pattern recognition}, pp.~8695--8704, 2020.

\bibitem{liu2020global}
Z.~Liu, X.~Qi, and P.~H. Torr, ``Global texture enhancement for fake face detection in the wild,'' in {\em Proceedings of the IEEE/CVF conference on computer vision and pattern recognition}, pp.~8060--8069, 2020.

\bibitem{hulzebosch2020detecting}
N.~Hulzebosch, S.~Ibrahimi, and M.~Worring, ``Detecting cnn-generated facial images in real-world scenarios,'' in {\em Proceedings of the IEEE/CVF conference on computer vision and pattern recognition workshops}, pp.~642--643, 2020.

\bibitem{guo2022robust}
H.~Guo, S.~Hu, X.~Wang, M.-C. Chang, and S.~Lyu, ``Robust attentive deep neural network for detecting gan-generated faces,'' {\em IEEE Access}, vol.~10, pp.~32574--32583, 2022.

\bibitem{pu2022learning}
W.~Pu, J.~Hu, X.~Wang, Y.~Li, S.~Hu, B.~Zhu, R.~Song, Q.~Song, X.~Wu, and S.~Lyu, ``Learning a deep dual-level network for robust deepfake detection,'' {\em Pattern Recognition}, vol.~130, p.~108832, 2022.

\bibitem{hu2021improving}
J.~Hu, S.~Wang, and X.~Li, ``Improving the generalization ability of deepfake detection via disentangled representation learning,'' in {\em 2021 IEEE International Conference on Image Processing (ICIP)}, pp.~3577--3581, IEEE, 2021.

\bibitem{zhang2020face}
K.-Y. Zhang, T.~Yao, J.~Zhang, Y.~Tai, S.~Ding, J.~Li, F.~Huang, H.~Song, and L.~Ma, ``Face anti-spoofing via disentangled representation learning,'' in {\em Computer Vision--ECCV 2020: 16th European Conference, Glasgow, UK, August 23--28, 2020, Proceedings, Part XIX 16}, pp.~641--657, Springer, 2020.

\bibitem{liang2022exploring}
J.~Liang, H.~Shi, and W.~Deng, ``Exploring disentangled content information for face forgery detection,'' in {\em European Conference on Computer Vision}, pp.~128--145, Springer, 2022.

\bibitem{Yan_2023_ICCV}
Z.~Yan, Y.~Zhang, Y.~Fan, and B.~Wu, ``Ucf: Uncovering common features for generalizable deepfake detection,'' in {\em Proceedings of the IEEE/CVF International Conference on Computer Vision (ICCV)}, pp.~22412--22423, October 2023.

\bibitem{zheng2024few}
P.~Zheng, H.~Chen, S.~Hu, B.~Zhu, J.~Hu, C.-S. Lin, X.~Wu, S.~Lyu, G.~Huang, and X.~Wang, ``Few-shot learning for misinformation detection based on contrastive models,'' {\em Electronics}, vol.~13, no.~4, p.~799, 2024.

\bibitem{chen2024masked}
T.~Chen, S.~Yang, S.~Hu, Z.~Fang, Y.~Fu, X.~Wu, and X.~Wang, ``Masked conditional diffusion model for enhancing deepfake detection,'' {\em arXiv preprint arXiv:2402.00541}, 2024.

\bibitem{lin2024detecting}
L.~Lin, N.~Gupta, Y.~Zhang, H.~Ren, C.-H. Liu, F.~Ding, X.~Wang, X.~Li, L.~Verdoliva, and S.~Hu, ``Detecting multimedia generated by large ai models: A survey,'' {\em arXiv preprint arXiv:2402.00045}, 2024.

\bibitem{fan2023synthesizing}
B.~Fan, S.~Hu, and F.~Ding, ``Synthesizing black-box anti-forensics deepfakes with high visual quality,'' {\em ICASSP}, 2024.

\bibitem{zhang2023x}
L.~Zhang, H.~Chen, S.~Hu, B.~Zhu, X.~Wu, J.~Hu, and X.~Wang, ``X-transfer: A transfer learning-based framework for robust gan-generated fake image detection,'' {\em arXiv preprint arXiv:2310.04639}, 2023.

\bibitem{yang2023improving}
S.~Yang, S.~Hu, B.~Zhu, Y.~Fu, S.~Lyu, X.~Wu, and X.~Wang, ``Improving cross-dataset deepfake detection with deep information decomposition,'' {\em arXiv preprint arXiv:2310.00359}, 2023.

\bibitem{fan2023attacking}
B.~Fan, Z.~Jiang, S.~Hu, and F.~Ding, ``Attacking identity semantics in deepfakes via deep feature fusion,'' in {\em 2023 IEEE 6th International Conference on Multimedia Information Processing and Retrieval (MIPR)}, pp.~114--119, IEEE, 2023.

\bibitem{chen2023harnessing}
H.~Chen, P.~Zheng, X.~Wang, S.~Hu, B.~Zhu, J.~Hu, X.~Wu, and S.~Lyu, ``Harnessing the power of text-image contrastive models for automatic detection of online misinformation,'' in {\em Proceedings of the IEEE/CVF Conference on Computer Vision and Pattern Recognition}, pp.~923--932, 2023.

\bibitem{trinh2021examination}
L.~Trinh and Y.~Liu, ``An examination of fairness of ai models for deepfake detection,'' {\em IJCAI}, 2021.

\bibitem{xu2022comprehensive}
Y.~Xu, P.~Terh{\"o}rst, K.~Raja, and M.~Pedersen, ``A comprehensive analysis of ai biases in deepfake detection with massively annotated databases,'' {\em arXiv preprint arXiv:2208.05845}, 2022.

\bibitem{news2}
K.~Wiggers, ``Deepfake detectors and datasets exhibit racial and gender bias, usc study shows,'' in {\em VentureBeat, \url{ https://tinyurl.com/ms8zbu6f}}, 2021.

\bibitem{nadimpalli2022gbdf}
A.~V. Nadimpalli and A.~Rattani, ``Gbdf: gender balanced deepfake dataset towards fair deepfake detection,'' {\em arXiv preprint arXiv:2207.10246}, 2022.

\bibitem{masood2022deepfakes}
M.~Masood, M.~Nawaz, K.~M. Malik, A.~Javed, A.~Irtaza, and H.~Malik, ``Deepfakes generation and detection: State-of-the-art, open challenges, countermeasures, and way forward,'' {\em Applied Intelligence}, pp.~1--53, 2022.

\bibitem{hazirbas2021towards}
C.~Hazirbas, J.~Bitton, B.~Dolhansky, J.~Pan, A.~Gordo, and C.~C. Ferrer, ``Towards measuring fairness in ai: the casual conversations dataset,'' {\em IEEE Transactions on Biometrics, Behavior, and Identity Science}, vol.~4, no.~3, pp.~324--332, 2021.

\bibitem{wang2022disentangled}
X.~Wang, H.~Chen, S.~Tang, Z.~Wu, and W.~Zhu, ``Disentangled representation learning,'' {\em arXiv preprint arXiv:2211.11695}, 2022.

\bibitem{pu2022fairness}
M.~Pu, M.~Y. Kuan, N.~T. Lim, C.~Y. Chong, and M.~K. Lim, ``Fairness evaluation in deepfake detection models using metamorphic testing,'' {\em arXiv preprint arXiv:2203.06825}, 2022.

\bibitem{rossler2019faceforensics++}
A.~Rossler, D.~Cozzolino, L.~Verdoliva, C.~Riess, J.~Thies, and M.~Nie{\ss}ner, ``Faceforensics++: Learning to detect manipulated facial images,'' in {\em Proceedings of the IEEE/CVF international conference on computer vision}, pp.~1--11, 2019.

\bibitem{googledeepfakes2019}
Google and Jigsaw, ``Deepfakes dataset by google \& jigsaw,'' in {\em \url{https://ai.googleblog.com/2019/09/contributing-data-to-deepfakedetection.html}}, 2019.

\bibitem{wang2023aleatoric}
H.~Wang, L.~He, R.~Gao, and F.~P. Calmon, ``Aleatoric and epistemic discrimination in classification,'' {\em ICML}, 2023.

\bibitem{wang2022understanding}
J.~Wang, X.~E. Wang, and Y.~Liu, ``Understanding instance-level impact of fairness constraints,'' in {\em International Conference on Machine Learning}, pp.~23114--23130, PMLR, 2022.

\bibitem{locatello2019fairness}
F.~Locatello, G.~Abbati, T.~Rainforth, S.~Bauer, B.~Sch{\"o}lkopf, and O.~Bachem, ``On the fairness of disentangled representations,'' {\em Advances in neural information processing systems}, vol.~32, 2019.

\bibitem{foret2020sharpness}
P.~Foret, A.~Kleiner, H.~Mobahi, and B.~Neyshabur, ``Sharpness-aware minimization for efficiently improving generalization,'' in {\em International Conference on Learning Representations}, 2020.

\bibitem{DeepFakes2017}
``Deepfakes,'' in {\em \url{https://github.com/deepfakes/faceswap}}, 2017.

\bibitem{thies2016face2face}
J.~Thies, M.~Zollhofer, M.~Stamminger, C.~Theobalt, and M.~Nie{\ss}ner, ``Face2face: Real-time face capture and reenactment of rgb videos,'' in {\em Proceedings of the IEEE conference on computer vision and pattern recognition}, pp.~2387--2395, 2016.

\bibitem{FaceSwap2018}
M.~Kowalski, ``Faceswap,'' in {\em \url{https://github.com/MarekKowalski/FaceSwap/}}, 2018.

\bibitem{thies2019deferred}
J.~Thies, M.~Zollh{\"o}fer, and M.~Nie{\ss}ner, ``Deferred neural rendering: Image synthesis using neural textures,'' {\em Acm Transactions on Graphics (TOG)}, vol.~38, no.~4, pp.~1--12, 2019.

\bibitem{li2019faceshifter}
L.~Li, J.~Bao, H.~Yang, D.~Chen, and F.~Wen, ``Faceshifter: Towards high fidelity and occlusion aware face swapping,'' {\em arXiv preprint arXiv:1912.13457}, pp.~2, 5, 2019.

\bibitem{mathews2023explainable}
S.~Mathews, S.~Trivedi, A.~House, S.~Povolny, and C.~Fralick, ``An explainable deepfake detection framework on a novel unconstrained dataset,'' {\em Complex \& Intelligent Systems}, pp.~1--13, 2023.

\bibitem{wang2019symmetric}
Y.~Wang, X.~Ma, Z.~Chen, Y.~Luo, J.~Yi, and J.~Bailey, ``Symmetric cross entropy for robust learning with noisy labels,'' in {\em Proceedings of the IEEE/CVF international conference on computer vision}, pp.~322--330, 2019.

\bibitem{cao2019learning}
K.~Cao, C.~Wei, A.~Gaidon, N.~Arechiga, and T.~Ma, ``Learning imbalanced datasets with label-distribution-aware margin loss,'' {\em Advances in neural information processing systems}, vol.~32, 2019.

\bibitem{oord2018representation}
A.~v.~d. Oord, Y.~Li, and O.~Vinyals, ``Representation learning with contrastive predictive coding,'' {\em arXiv preprint arXiv:1807.03748}, 2018.

\bibitem{huang2017arbitrary}
X.~Huang and S.~Belongie, ``Arbitrary style transfer in real-time with adaptive instance normalization,'' in {\em Proceedings of the IEEE international conference on computer vision}, pp.~1501--1510, 2017.

\bibitem{dwork2012fairness}
C.~Dwork, M.~Hardt, T.~Pitassi, O.~Reingold, and R.~Zemel, ``Fairness through awareness,'' in {\em Proceedings of the 3rd innovations in theoretical computer science conference}, pp.~214--226, 2012.

\bibitem{hardt2016equality}
M.~Hardt, E.~Price, and N.~Srebro, ``Equality of opportunity in supervised learning,'' {\em Advances in neural information processing systems}, vol.~29, 2016.

\bibitem{hu2023rank}
S.~Hu, X.~Wang, and S.~Lyu, ``Rank-based decomposable losses in machine learning: A survey,'' {\em IEEE Transactions on Pattern Analysis and Machine Intelligence}, 2023.

\bibitem{hu2022distributionally}
S.~Hu and G.~H. Chen, ``Distributionally robust survival analysis: A novel fairness loss without demographics,'' in {\em Machine Learning for Health}, pp.~62--87, PMLR, 2022.

\bibitem{hu2022sum}
S.~Hu, Y.~Ying, X.~Wang, and S.~Lyu, ``Sum of ranked range loss for supervised learning,'' {\em The Journal of Machine Learning Research}, vol.~23, no.~1, pp.~4826--4869, 2022.

\bibitem{hu2021tkml}
S.~Hu, L.~Ke, X.~Wang, and S.~Lyu, ``Tkml-ap: Adversarial attacks to top-k multi-label learning,'' in {\em Proceedings of the IEEE/CVF International Conference on Computer Vision}, pp.~7649--7657, 2021.

\bibitem{hu2020learning}
S.~Hu, Y.~Ying, S.~Lyu, {\em et~al.}, ``Learning by minimizing the sum of ranked range,'' {\em Advances in Neural Information Processing Systems}, vol.~33, pp.~21013--21023, 2020.

\bibitem{hu2023outlier}
S.~Hu, Z.~Yang, X.~Wang, Y.~Ying, and S.~Lyu, ``Outlier robust adversarial training,'' {\em ACML}, 2023.

\bibitem{williamson2019fairness}
R.~Williamson and A.~Menon, ``Fairness risk measures,'' in {\em International Conference on Machine Learning}, pp.~6786--6797, PMLR, 2019.

\bibitem{levy2020large}
D.~Levy, Y.~Carmon, J.~C. Duchi, and A.~Sidford, ``Large-scale methods for distributionally robust optimization,'' {\em Advances in Neural Information Processing Systems}, vol.~33, pp.~8847--8860, 2020.

\bibitem{deepfakedetection2021}
``Deepfake detection challenge.'' \url{https://www.kaggle.com/c/deepfake-detection-challenge}.
\newblock Accessed: 2021-04-24.

\bibitem{li2020celebdf}
Y.~Li, X.~Yang, P.~Sun, H.~Qi, and S.~Lyu, ``Celeb-df: A new dataset for deepfake forensics,'' in {\em CVPR}, pp.~6,7, 2020.

\bibitem{luo2021generalizing}
Y.~Luo, Y.~Zhang, J.~Yan, and W.~Liu, ``Generalizing face forgery detection with high-frequency features,'' in {\em CVPR}, 2021.

\bibitem{chollet2017xception}
F.~Chollet, ``Xception: Deep learning with depthwise separable convolutions,'' in {\em Proceedings of the IEEE conference on computer vision and pattern recognition}, pp.~1251--1258, 2017.

\bibitem{he2016deep}
K.~He, X.~Zhang, S.~Ren, and J.~Sun, ``Deep residual learning for image recognition,'' in {\em Proceedings of the IEEE conference on computer vision and pattern recognition}, pp.~770--778, 2016.

\bibitem{tan2019efficientnet}
M.~Tan and Q.~Le, ``Efficientnet: Rethinking model scaling for convolutional neural networks,'' in {\em International conference on machine learning}, pp.~6105--6114, PMLR, 2019.

\bibitem{selvaraju2017gradcam}
R.~R. Selvaraju, M.~Cogswell, A.~Das, R.~Vedantam, D.~Parikh, and D.~Batra, ``Grad-cam: Visual explanations from deep networks via gradient-based localization,'' in {\em Proceedings of the IEEE international conference on computer vision}, pp.~618--626, 2017.

\bibitem{hu2021exposing}
S.~Hu, Y.~Li, and S.~Lyu, ``Exposing gan-generated faces using inconsistent corneal specular highlights,'' in {\em ICASSP 2021-2021 IEEE International Conference on Acoustics, Speech and Signal Processing (ICASSP)}, pp.~2500--2504, IEEE, 2021.

\bibitem{guo2022eyes}
H.~Guo, S.~Hu, X.~Wang, M.-C. Chang, and S.~Lyu, ``Eyes tell all: Irregular pupil shapes reveal gan-generated faces,'' in {\em ICASSP 2022-2022 IEEE International Conference on Acoustics, Speech and Signal Processing (ICASSP)}, pp.~2904--2908, IEEE, 2022.

\bibitem{guo2022open}
H.~Guo, S.~Hu, X.~Wang, M.-C. Chang, and S.~Lyu, ``Open-eye: An open platform to study human performance on identifying ai-synthesized faces,'' in {\em 2022 IEEE 5th International Conference on Multimedia Information Processing and Retrieval (MIPR)}, pp.~224--227, IEEE, 2022.

\bibitem{li2018ictu}
Y.~Li, M.-C. Chang, and S.~Lyu, ``In ictu oculi: Exposing ai created fake videos by detecting eye blinking,'' in {\em 2018 IEEE International workshop on information forensics and security (WIFS)}, pp.~1--7, IEEE, 2018.

\bibitem{matern2019exploiting}
F.~Matern, C.~Riess, and M.~Stamminger, ``Exploiting visual artifacts to expose deepfakes and face manipulations,'' in {\em 2019 IEEE Winter Applications of Computer Vision Workshops (WACVW)}, pp.~83--92, IEEE, 2019.

\bibitem{yang2019exposing}
X.~Yang, Y.~Li, H.~Qi, and S.~Lyu, ``Exposing gan-synthesized faces using landmark locations,'' in {\em Proceedings of the ACM workshop on information hiding and multimedia security}, pp.~113--118, 2019.

\bibitem{qian2020thinking}
Y.~Qian, G.~Yin, L.~Sheng, Z.~Chen, and J.~Shao, ``Thinking in frequency: Face forgery detection by mining frequency-aware clues,'' in {\em European conference on computer vision}, pp.~86--103, Springer, 2020.

\bibitem{khayatkhoei2022spatial}
M.~Khayatkhoei and A.~Elgammal, ``Spatial frequency bias in convolutional generative adversarial networks,'' in {\em Proceedings of the AAAI Conference on Artificial Intelligence}, vol.~36, pp.~7152--7159, 2022.

\bibitem{dzanic2020fourier}
T.~Dzanic, K.~Shah, and F.~Witherden, ``Fourier spectrum discrepancies in deep network generated images,'' {\em Advances in neural information processing systems}, vol.~33, pp.~3022--3032, 2020.

\bibitem{frank2020leveraging}
J.~Frank, T.~Eisenhofer, L.~Sch{\"o}nherr, A.~Fischer, D.~Kolossa, and T.~Holz, ``Leveraging frequency analysis for deep fake image recognition,'' in {\em International conference on machine learning}, pp.~3247--3258, PMLR, 2020.

\bibitem{zhang2019detecting}
X.~Zhang, S.~Karaman, and S.-F. Chang, ``Detecting and simulating artifacts in gan fake images,'' in {\em 2019 IEEE international workshop on information forensics and security (WIFS)}, pp.~1--6, IEEE, 2019.

\bibitem{mcinnes2018umap}
L.~McInnes, J.~Healy, and J.~Melville, ``Umap: Uniform manifold approximation and projection for dimension reduction,'' {\em arXiv preprint arXiv:1802.03426}, 2018.

\end{thebibliography}
